\documentclass[journal]{IEEEtran}
\usepackage{cite}
\usepackage{amsmath,amssymb,amsfonts}
\DeclareMathOperator*{\argmin}{argmin} % no space, limits underneath in displays
\usepackage{algorithmic}
\usepackage{graphicx}
\usepackage{textcomp}
\usepackage{threeparttable} %http://blog.modelworks.ch/tables-with-footnotes-in-latex/
\usepackage{footnote}
\usepackage{multirow} %https://blog.csdn.net/lishoubox/article/details/7330409
\usepackage{pifont}% http://ctan.org/pkg/pifont
\usepackage{times}
\usepackage{color, colortbl}  %the first to define new colors and the latter to actually color the table
\usepackage[table]{xcolor}
\usepackage{graphicx}
\usepackage{bbding}
\usepackage{soul}
\usepackage[utf8]{inputenc}
\usepackage{overpic}
\usepackage{booktabs}
\usepackage{diagbox}
\usepackage{etoolbox}
\usepackage{url}
\usepackage{rotating}         %用于旋转对象 https://blog.csdn.net/lqhbupt/article/details/46586709
\apptocmd{\sloppy}{\hbadness 10000\relax}{}{}
\usepackage[multiple]{footmisc}
\definecolor{peach-orange}{rgb}{1,0.8,0.6}
\definecolor{light_blue}{rgb}{0.6,0.81,0.93}
\usepackage[colorlinks=true,pdftex]{hyperref} %\usepackage[hidelinks]{hyperref}  %
\newcommand{\figref}[1]{Fig.~\ref{#1}}%
\newcommand{\tabref}[1]{Tab.~\ref{#1}}%
\newcommand{\secref}[1]{Sec.~\ref{#1}}
\renewcommand{\eqref}[1]{Eq.~(\ref{#1})}
\newcommand{\eg}[0]{{\emph{e.g.}}}% 例如
\newcommand{\etal}[0]{{\emph{et. al.}}}% 把主要作者列出后，其它作者全放在et al. 里面， 人的场合用et al，而无生命的场合用etc.
\newcommand{\ie}[0]{{i.e.}}% 也就是说,即
\newcommand{\etc}[0]{{etc}}% 等等
\newcommand{\tabincell}[1]{\begin{tabular}{@{}c@{}}#1\end{tabular}}   %replace c with #1 for alignment control
\def\BibTeX{{\rm B\kern-.05em{\sc i\kern-.025em b}\kern-.08em
    T\kern-.1667em\lower.7ex\hbox{E}\kern-.125emX}}
\begin{document}
%\title{IrisParseNet: A Novel Deep Multi-task Learning Framework for Iris Segmentation and Localization}  %IrisParsingNet
\title{Joint Iris Segmentation and Localization Using Deep Multi-task Learning Framework}  %IrisParsingNet
\author{Caiyong Wang, Yuhao Zhu, Yunfan Liu, Ran He, \IEEEmembership{Senior Member, IEEE}, and Zhenan Sun, \IEEEmembership{Member, IEEE}
%\thanks{Caiyong Wang, Yuhao Zhu, Yunfan Liu, Ran He and Zhenan Sun (Corresponding author) are with the Center for
%Research on Intelligent Perception and Computing, National Laboratory of Pattern Recognition, Institute of Automation, Chinese Academy of Sciences,
%Beijing 100190, China. E-mail: \{caiyong.wang, yuhao.zhu, yunfan.liu\}@cripac.ia.ac.cn, \{rhe, znsun\}@nlpr.ia.ac.cn.}}
%\thanks{ This work was supported by the National Natural Science Foundation of China (Grant No. 61427811, 61573360) and the National Key Research and Development Program of China (Grant No. 2017YFC0821602, 2016YFB1001000). (Corresponding author: Zhenan Sun.)}
\thanks{The supplemental material is made available via \url{https://drive.google.com/open?id=1Fo3Nmv_ha5-d5jC2vcbAtbjMyJ_aa_fL}}
\thanks{Caiyong Wang is with the School of Artificial Intelligence, University of Chinese
Academy of Sciences, Beijing 100049, China, and also with the Center for
Research on Intelligent Perception and Computing, National Laboratory of
Pattern Recognition, Institute of Automation, Chinese Academy of Sciences,
Beijing 100190, China. E-mail: wangcaiyong2017@ia.ac.cn.}
%\thanks{Yuhao Zhu, Yunfan Liu, Ran He and Zhenan Sun(\textbf{Corresponding author}) are with the Center for
%Research on Intelligent Perception and Computing, National Laboratory of
%Pattern Recognition, Institute of Automation, Chinese Academy of Sciences,
%Beijing 100190, China. Ran He and Zhenan Sun are also with the CAS Center for
%Excellence in Brain Science and Intelligence Technology, Chinese Academy
%of Sciences, Beijing 100190, China. E-mail: \{yuhao.zhu, yunfan.liu\}@cripac.ia.ac.cn, \{rhe, znsun\}@nlpr.ia.ac.cn.}}
\thanks{Yuhao Zhu, Yunfan Liu, Ran He and Zhenan Sun(\textbf{Corresponding author}) are with the Center for
Research on Intelligent Perception and Computing, National Laboratory of
Pattern Recognition, Institute of Automation, Chinese Academy of Sciences,
Beijing 100190, China. E-mail: \{yuhao.zhu, yunfan.liu\}@cripac.ia.ac.cn, \{rhe, znsun\}@nlpr.ia.ac.cn.}}

\maketitle

\begin{abstract}
Iris segmentation and localization in non-cooperative environment is challenging due to illumination variations, long distances, moving subjects and limited user cooperation,~\etc. Traditional methods often suffer from poor performance when confronted with iris images captured in these conditions. Recent studies have shown that deep learning methods could achieve impressive performance on
iris segmentation task\cite{liu2016accurate,Jalilian2017Domain,Bazrafkan2018An,severo2018benchmark,arsalan2018irisdensenet}. In addition, as iris is defined as an annular region between pupil and sclera, geometric constraints could be imposed to help locating the iris more accurately and improve the segmentation results.
In this paper, we propose a deep multi-task learning framework, named as IrisParseNet, to exploit the inherent correlations between pupil, iris and sclera to boost up the performance of iris segmentation and localization in a unified model.
In particular, IrisParseNet firstly applies a Fully Convolutional Encoder-Decoder Attention Network to simultaneously estimate pupil center, iris segmentation mask and iris inner/outer boundary. Then, an effective post-processing method is adopted for iris inner/outer circle localization.
To train and evaluate the proposed method, we manually label three challenging iris datasets, namely CASIA-Iris-Distance, UBIRIS.v2, and MICHE-I, which cover various types of noises.
Extensive experiments are conducted on these newly annotated datasets, and results show that our method outperforms state-of-the-art methods on various benchmarks.
All the ground-truth annotations, annotation codes and evaluation protocols are publicly available at \url{https://github.com/xiamenwcy/IrisParseNet}.
\end{abstract}


\section{Introduction}
\label{sec:introduction}
\IEEEPARstart{I}{r}is recognition has been considered as one of the most stable, accurate and reliable biometric identification technologies\cite{jain2007handbook}, hence it is widely applied in various biometric applications including intelligent unlocking, border crossing control, security and crime screening,~\etc.
A complete iris recognition system often consists of four sub-processes: iris image acquisition, iris preprocessing, feature extraction and matching. Both Iris segmentation and iris inner/outer circle localization (iris localization) are part of the iris preprocessing step\cite{wildes1997iris,daugman2009iris}.
\begin{figure}[!t]
  \begin{overpic}[width=1.\linewidth]{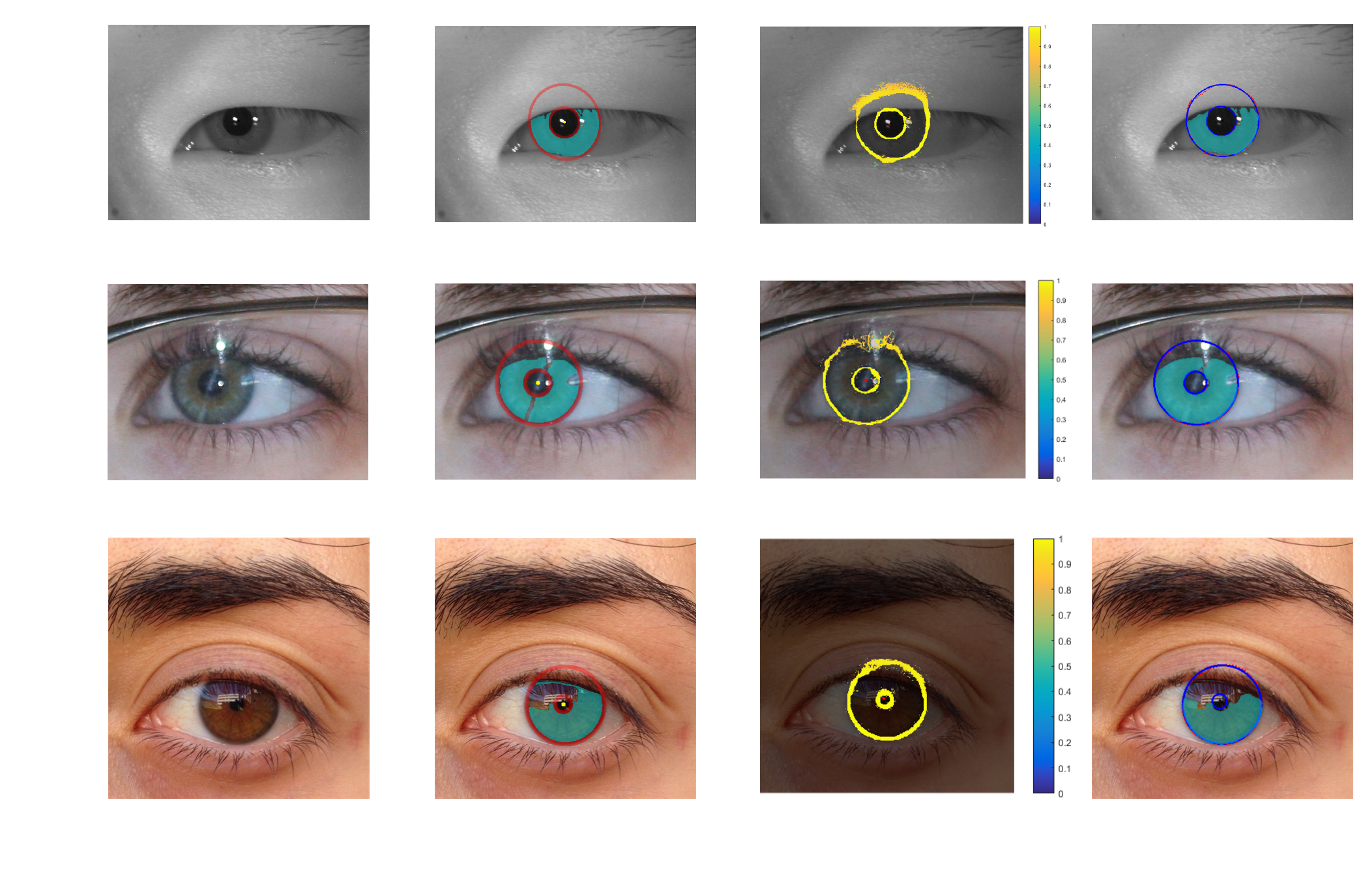}
  \put(15,1){(a)}
  \put(38,1){(b)}
  \put(63,1){(c)}
  \put(86,1){(d)}
  \put(3,52){\begin{sideways}{\scriptsize CASIA}\end{sideways}} %CASIA-Iris-Distance
  \put(3,33){\begin{sideways}{\scriptsize UBIRIS}\end{sideways}} %UBIRIS.v2
  \put(3,12){\begin{sideways}{\scriptsize MICHE}\end{sideways}} % MICHE-I
  \end{overpic}
  \vspace{-13pt}
  \caption{
  The first column (a) shows iris images from three datasets (as described in \secref{sec::dataset}) collected in different environments.
  The second column (b) illustrates the ground truths of pupil center, iris inner/outer boundary and iris segmentation mask, highlighted in yellow, red and aqua, respectively.
  The third column (c) shows the predicted pupil center (marked as red) and iris inner/outer boundary (highlighted in a color bar where the hotter color indicates the higher probability of a pixel belonging to the actual iris boundary). By utilizing the inherent correlation of pupil center, iris mask (highlighted in aqua in the column (d)) and iris inner/outer boundary, we further eliminate the noise of detected iris boundaries. As shown in the fourth column (d), with the help of refined iris boundaries and pupil center, we could extract coarse iris contours (highlighted in red) as the fitting points, then locate iris inner/outer circle (highlighted in blue) with the least-squares circle fitting algorithm\cite{chernov2005least}. Best viewed in color.}
  \label{fig:iris_introd}
  \vspace{-10pt}
\end{figure}
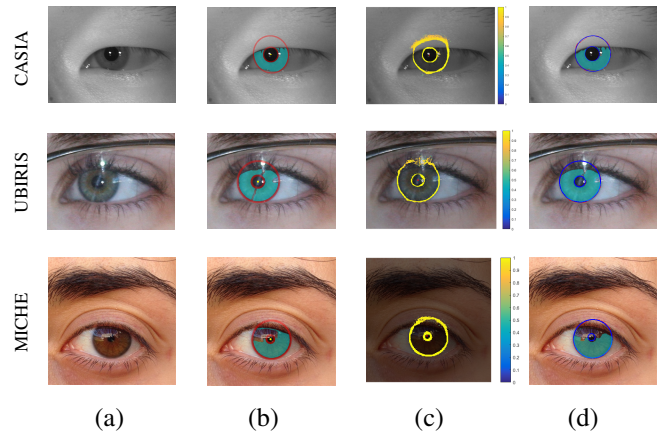

As shown in \figref{fig:iris_introd} (a) and (b), iris refers to an annular region between pupil and sclera. Iris boundaries are approximately
defined by two circles,~\ie~an inner circle that divides pupil and iris (also called pupillary boundary), and an outer circle that separates iris and sclera (also called limbic boundary).
Iris segmentation aims to isolate valid iris texture region from other components, such as pupil, sclera, eyelashes, eyelids, reflections, and occlusions
in an eye image to obtain a binary mask, where valid iris pixels are classified as foreground and other pixels are regarded as background.
Iris localization refers to estimating the parameters (center and radius) of iris inner and outer circular boundaries.
After obtaining parameters of the iris region, normalization is carried out to get normalized image and mask, then followed by feature extraction and match operations to produce the final recognition result.
As the beginning of iris recognition flow, accurate segmentation and localization has a great impact on subsequent processes\cite{Hofbauer2016Experimental,Proen2010Iris}. Therefore, a segmentation and localization algorithm with high performance is the key to the success of the entire iris recognition system.

\begin{figure}[ht]
  \begin{overpic}[width=1\linewidth]{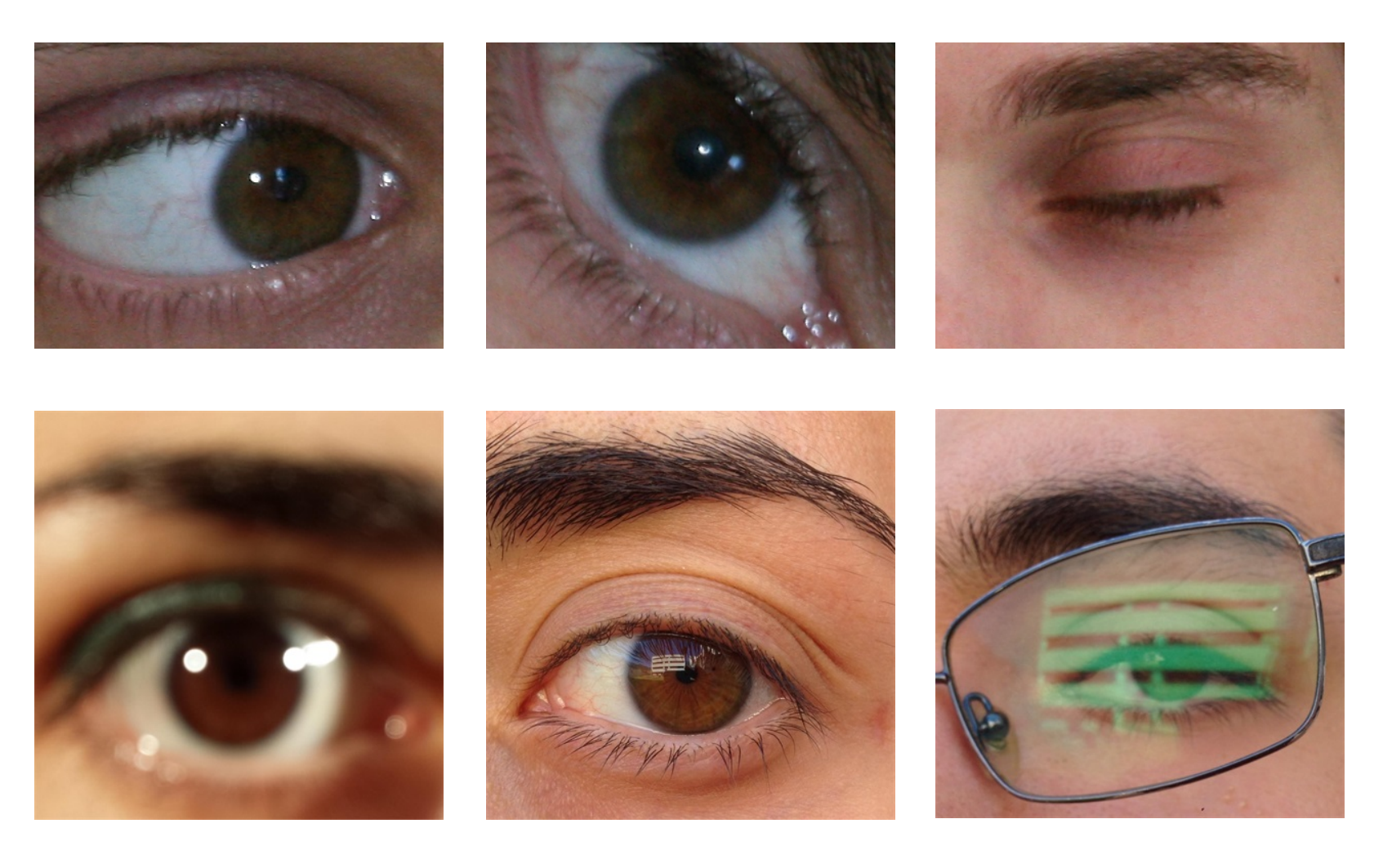}
  \put (15,34) {(a)}
  \put (47,34) {(b)}
  \put (80,34) {(c)}
  \put (15,-1) {(d)}
  \put (47,-1) {(e)}
  \put (80,-1) {(f)}
  \end{overpic}\vspace{-4pt}
  \caption{Examples of degraded iris images with different types of noises. (a) gaze deviation; (b) rotation images;
  (c) absence of iris; (d) defocus blur; (e) specular reflections; (f) iris occlusions due to glasses.}
  \label{fig:iris_noise}
\end{figure}
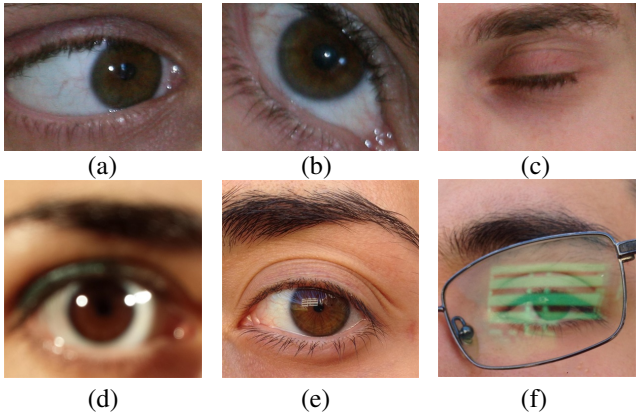

Earlier iris recognition systems require user cooperation and highly controlled imaging conditions, which restricts the applications of iris recognition technology. Hence, it is necessary to develop less constrained iris recognition systems.
However, images captured in less constrained scenarios (\eg~long distances, moving subjects, using mobile devices, and limited user cooperation) are often of poor quality and introduce various kinds of noise, such as partial occlusions due to eyelids or glasses and blur caused by motion and defocus, as shown in \figref{fig:iris_noise}.

Over the past decades, a number of methods have been proposed for iris segmentation and localization, such as Hough transform \cite{wildes1997iris,Zhao2015An}, Active Contours\cite{Banerjee2016}, and GrowCut\cite{RADMAN201760}.
However, these methods could not work well when dealing with degraded images.
Compared with these traditional approaches, deep learning models, especially Convolutional Neural Networks (CNNs), have shown incomparable advantages in tasks such as image classification\cite{krizhevsky2012imagenet} and object detection\cite{ren2015faster}.
To be specific, hierachical semantic representations of the input image could be automatically learned in an end-to-end manner without requiring extra human efforts.
Since the rapid development of deep learning, a large amount of studies using CNNs have been proposed for iris segmentation\cite{liu2016accurate,Jalilian2017Domain,Bazrafkan2018An,severo2018benchmark,arsalan2018irisdensenet}, iris bounding box detection\cite{severo2018benchmark}, and pupil center detection\cite{chinsatit2017cnn,veraolmos2017deconvolutional,park2018learning}.
However, to the best of our knowledge, little research attention has been devoted to locating iris inner and outer boundaries based on deep learning technology. In addition, the geometric structure of iris,~\ie~the pupil center is inside the inner boundary of the iris and the iris mask is located in between the inner and outer boundaries of the iris, could serve as priori constraints in designing iris segmentation and localization algorithms.

Based on these observations, we propose a deep multi-task learning framework for simultaneous pupil center detection, iris segmentation and iris inner/outer boundary detection, followed by an effective post-processing operation for iris localization, as shown in \figref{fig:iris_introd} (c) and (d).
Compared with single objective learning, joint learning of multi-modal eye structures makes the network learn more discriminative and essential features.

To train and evaluate the proposed model, we collect three challenging public iris datasets: CASIA-Iris-Distance\cite{casiav4}, UBIRIS.v2\cite{UBIRISv2} and MICHE-I\cite{miche_dataset}. All these datasets contain segmentation annotations provided by other literatures.
We also manually label pupil center and iris inner/outer boundary as additional ground truths for each iris image. These datasets contain various categories of noises such as blur, off-axis, occlusions and specular reflections, which could evaluate the robustness of the proposed method.
To promote the research on iris preprocessing, we have made our manually annotated labels freely available to the community.

Main contributions of this paper are summarized as follows:

\begin{enumerate}
\item  This paper introduces a novel multi-task framework which consists of two parts: the first part is a Fully Convolutional Encoder-Decoder Network equipped with attention modules which could learn more discriminative features for producing multiple probability maps. By optimizing focal loss\cite{lin2017focal} and balanced sigmoid cross-entropy loss\cite{xie2015hed}, the model could alleviate the class-imbalanced problem and converge quickly. The second part is an effective post-processing method including edge denoising, Viterbi-based coarse contours detection\cite{Sutra2012The} and least-squares circle fitting\cite{chernov2005least} for iris localization.
\item  We select three representative iris datasets and label the pupil center as well as inner/outer boundary for each iris image. Furthermore, we build comprehensive evaluation protocols for evaluating the performance of iris segmentation and localization algorithms.
\item  The proposed method achieves state-of-the-art results on various iris benchmarks. Moreover, it has strong robustness and generalization ability, providing a good foundation for subsequent iris recognition processes.
\end{enumerate}

The  paper is organized as follows. In Section II, we briefly review related work on iris segmentation and iris localization. Technical details of the proposed method are elaborated in Section III. Section IV introduces three databases and the annotation method that we adopt. Section V describes the evaluation protocols and analyzes experimental results. Finally, we conclude the paper and discuss future work in Section VI.

\section{Related Work}
This section provides an overview of literatures on iris segmentation, semantic edge detection and iris localization.

\subsection{Iris Segmentation}
\label{segsection}
Over the past decades, a number of methods are proposed for iris segmentation. In general, these segmentation methods could be classified into two main categories: boundary-based methods and pixel-based methods\cite{liu2016accurate}. Boundary-based methods mainly locate pupillary, limbic and eyelid boundaries to isolate iris texture regions. On the contrary, pixel-based methods directly distinguish iris pixels from non-iris pixels according to the pixel-level appearance information.

For boundary-based methods, Daugman's integro-differential operator\cite{Daugman_id} and Wilde's circular Hough transforms\cite{wildes1997iris} are the two most well-known algorithms. The most critical and fundamental assumption these two methods made is that pupillary and limbic boundaries are circular contours. The integro-differential operator searches for the largest difference of intensity over the parameter space which normally corresponds to pupil and iris boundaries, while Hough transforms find optimal curve parameters by a voting procedure in a binary edge image. Although these methods have achieved good segmentation performance in iris images captured in controlled environments, they are time consuming and not suitable for degraded iris images. To overcome these problems, many noise removal\cite{Zhao2015An}, coarse iris location\cite{haindl2015unsupervised,gangwar2016irisseg} and multiple models selection\cite{HU201524} methods have been proposed to improve the robustness and efficiency of bounding-based iris segmentation methods. Besides, since the pupil and iris boundaries are not strictly circular, some works attempted to use geodesic active contours \cite{shah2009iris} or elliptic contours \cite{Banerjee2016} to replace the circular assumption.

On the other hand, pixel-based methods exploit low-level visual information of individual pixel, such as intensity and color, to classify the pixels of interest from the background of the image. The most promising method in this category use commonly known pixel-level techniques, such as Graph Cut\cite{RADMAN201760,Banerjee2016}, to pre-process the image and traditional classification methods, such as SVMs\cite{rongnian2011improving}, to classify the iris pixels from non-iris pixels.

Current boundary-based and pixel-based methods are designed mainly based on prior knowledge and require much pre- and post-processing effort. Deep learning models, especially Convolutional Neural Networks (CNNs), provide a powerful end-to-end solution to effectively solve these problems.

Semantic segmentation could be considered as a pixel-wise image classification task, \ie~each pixel in the image is assigned an object class. In 2005, Long\cite{long2015fully} \etal~firstly proposed Fully Convolutional Network (FCN) for semantic segmentation. Afterwards, a number of semantic segmentation methods based on FCN have been proposed, such as DeepLab series\cite{chen2014semantic,chen2016deeplab,chen2017rethinking}, U-Net\cite{ronneberger2015u-net}, and PSPNet\cite{zhao2017pyramid} to improve the performance of semantic segmentation. FCN-based methods take the whole image as input and produce a probability density map through a series of convolutional layers without involving fully connected layers. The whole model is end-to-end, which does not require any manual processing, and could achieve state-of-the-art performances of the time. Iris segmentation could be regarded as a special binary semantic segmentation problem. Hence, many FCN-based segmentation methods could be directly applied on iris images, such as \cite{liu2016accurate,Jalilian2017Domain,Bazrafkan2018An,arsalan2018irisdensenet}.
Inspired by the success of U-Net on binary semantic segmentation task\cite{LinkNet,TernausNet,Robot2018}, in this paper, we propose a Fully Convolutional Encoder-Decoder Attention Network for iris segmentation.

\subsection{Semantic Edge Detection \& Iris Localization}
Edge detection is a classical challenge in computer vision. Previous to the rapid development of deep learning, well-known Sobel detctor and Canny detector\cite{canny1986computational} \etc~are widely adopted. However, traditional methods are difficult to deal with semantic edges, \ie~edges which we are interseted in. Therefore, a lot of deep learning based methods\cite{xie2015hed,yu2017casenet} are proposed to solve the semantic edge detection problem. Most of these methods adopt Fully Convolutional Networks (FCNs) and directly concatenate the features of different stages to extract semantic edges. In this paper, we mainly concentrate on iris inner/outer boundary detection using deep learning models.

Classical iris localization methods usually involve Daugman's integral differential operator\cite{Daugman_id}, Wildes's circular Hough Transform\cite{wildes1997iris} and their variants, as described in \secref{segsection}. The main idea of these methods is directly searching for the optimal parameters of inner and outer circular boundaries of iris in the parameter space. These methods are efficient but only suitable for iris images without severe distortions and noises. Different from these methods, edge detection based iris localization methods have demonstrated their superiorities on non-ideal iris images. In \cite{gangwar2016irisseg}, the author adopted coarse-to-fine strategy to localize inner and outer boundaries of iris. Inner boundary is coarsely detected using an iterative search method by exploiting dynamic thresholds and multiple local cues, and outer boundary is first approximated in polar space using adaptive filters, then refined in the cartesian space. As a result, these two boundaries are robust against noises and distortions in iris images, which facilitates the subsequent circle fitting process. In \cite{Sutra2012The}, the Viterbi algorithm is applied on gradient maps of iris images to find coarse low-resolution contours which means selecting the least number of noisy gradients points as possible, then followed by least-squares circle fitting\cite{chernov2005least} for iris localization. Experiment results indicate that the method is accurate and robust, and does not require refined parameter adaptation to various degradations encountered. In this paper, we adopt the method proposed in \cite{Sutra2012The} as the main body of our post-processing step, and use real iris boundaries extracted by deep learning models in replace of gradient maps.

\section{Technical Details}
In this section, we firstly introduce the whole pipeline of our method. After that, we elaborate on the proposed multi-task network framework based on Fully Convolutional Network and attention mechanism, followed by an effective post-processing approach. Finally we describe our training objectives of the proposed model.
\subsection{Pipeline}
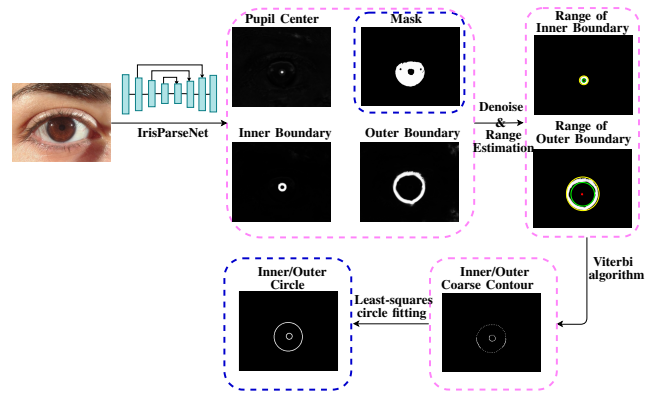
\begin{figure}[!htb]
  \centering
  \begin{overpic}[width=0.95\linewidth]{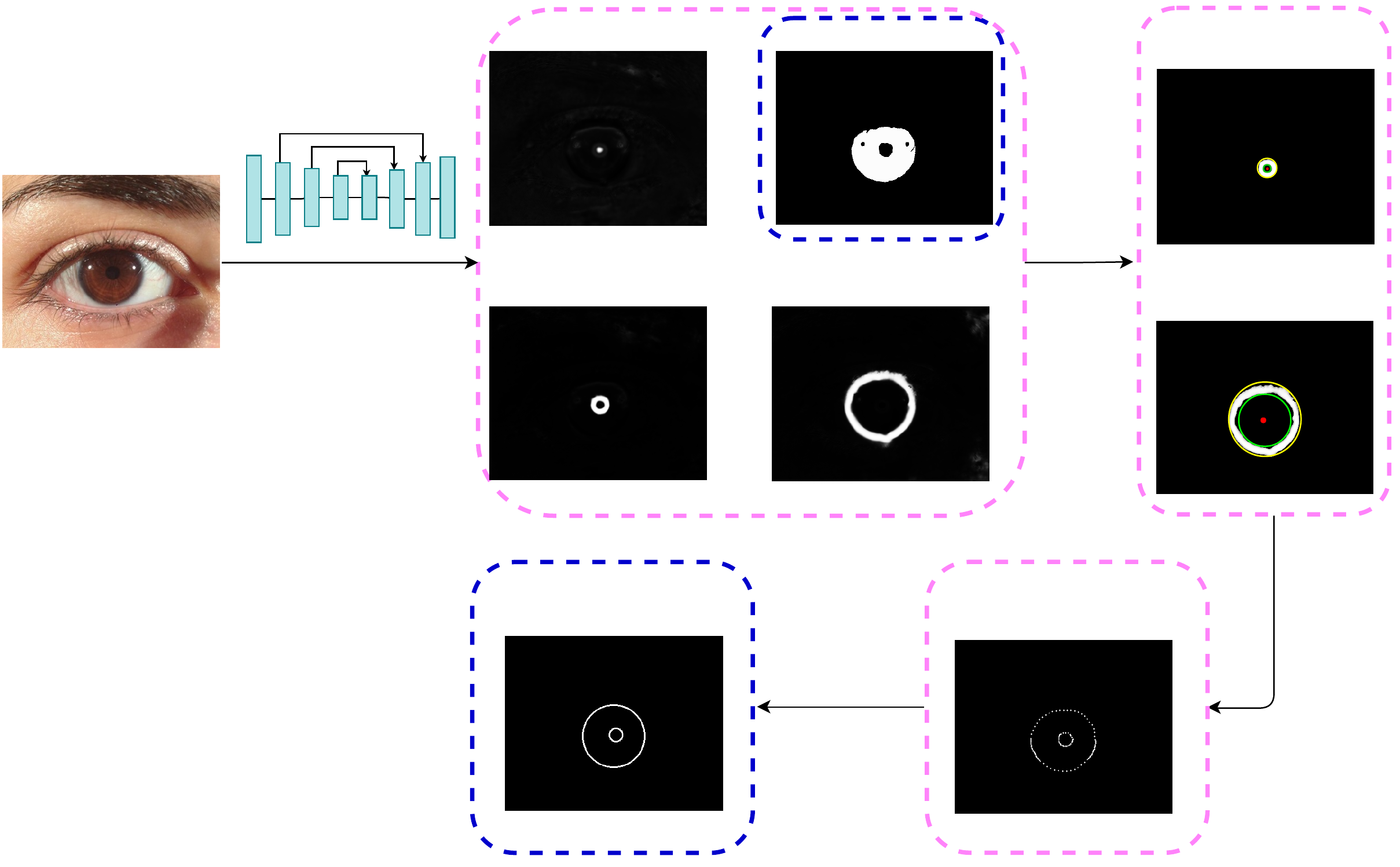}
  \put(20,41){\tiny \textbf{IrisParseNet}}
     \put(37,59){\tiny \textbf{Pupil Center}}
     \put(60,59){\tiny \textbf{Mask}}
     \put(36,41){\tiny \textbf{Inner Boundary}}
     \put(56,41){\tiny \textbf{Outer Boundary}}
       \put(74,45){\tiny \textbf{Denoise}}
     \put(76.5,42.5){\tiny \textbf{\&}}
     \put(75,40.6){\tiny \textbf{Range}}
     \put(73,38.6){\tiny \textbf{Estimation}}
     \put(86,59.5){\tiny \textbf{Range of}}
     \put(83,57.5){\tiny \textbf{Inner Boundary}}
      \put(86,42){\tiny \textbf{Range of}}
     \put(83,40){\tiny \textbf{Outer Boundary}}
     \put(92,20.5){\tiny \textbf{Viterbi}}
     \put(91.3,18.5){\tiny \textbf{algorithm}}
     \put(71,19){\tiny \textbf{Inner/Outer}}
     \put(68,17){\tiny \textbf{Coarse Contour}}
     \put(39,19){\tiny \textbf{Inner/Outer}}
     \put(41,17){\tiny \textbf{Circle}}
     \put(54.3,14.5){\tiny \textbf{Least-squares}}
     \put(54.7,12.5){\tiny \textbf{circle fitting}}
  \end{overpic}
  \vspace{-8pt}
  \caption{The pipeline of proposed method: network output and post-processing.}
  \label{fig:pipline}
\end{figure}
The pipeline of the proposed method is illustrated in \figref{fig:pipline}. IrisParseNet predicts probability maps of pupil center, iris segmentation mask and iris inner/outer boundary. Then, we further utilize the prior geometry relations of these elements to exclude mispredicted results, remove outliers and get the range of iris inner/outer circular boundary(\ie~circle center, minimum/maximum radius). Subsequently, Viterbi algorithm\cite{Sutra2012The} is used to extract coarse iris inner/outer contour. Finally iris inner/outer circle is localized by fitting on these coarse iris contours.

\subsection{Multi-task Network Framework}
\begin{figure*}[!htb]
\centering
  \begin{overpic}[width=\linewidth]{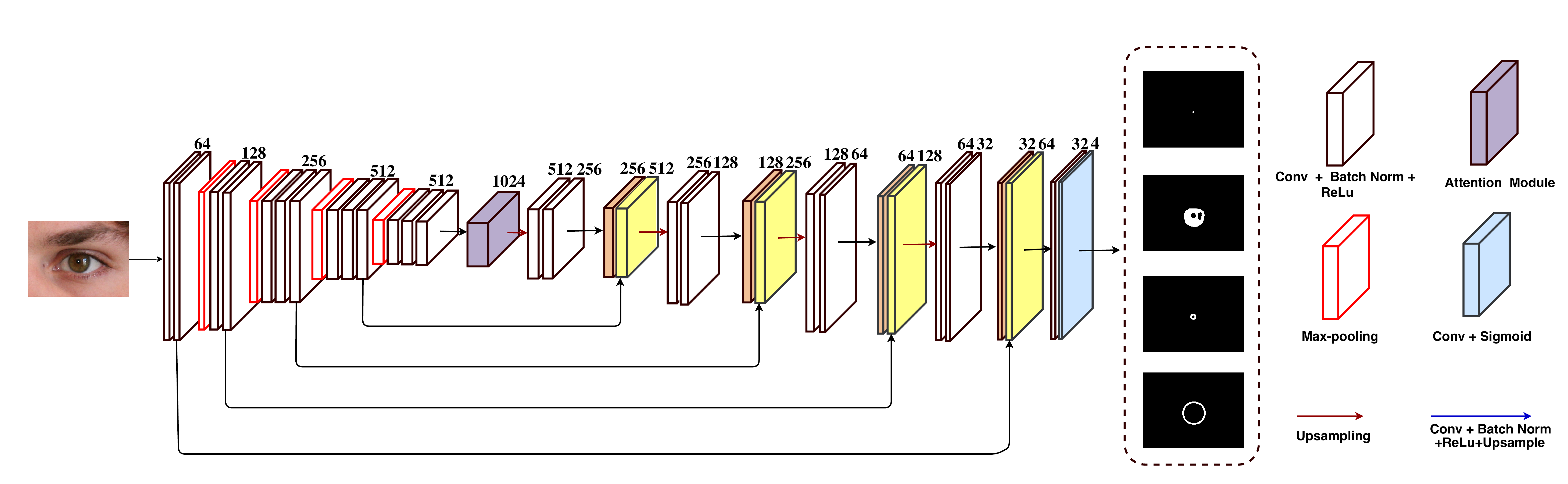}
    \put(73.5,26.6){\tiny \textbf{Pupil Center}}
    \put(75,20){\tiny \textbf{Mask}}
    \put(72.5,13.5){\tiny \textbf{Inner Boundary}}
    \put(72.5,7.5){\tiny \textbf{Outer Boundary}}
  \end{overpic}
  %\vspace{6pt}
  \caption{Overview of Multi-task Attention Network Architecture. Best viewed in color.}
  \label{fig:arch-network}\vspace{-10pt}
\end{figure*}
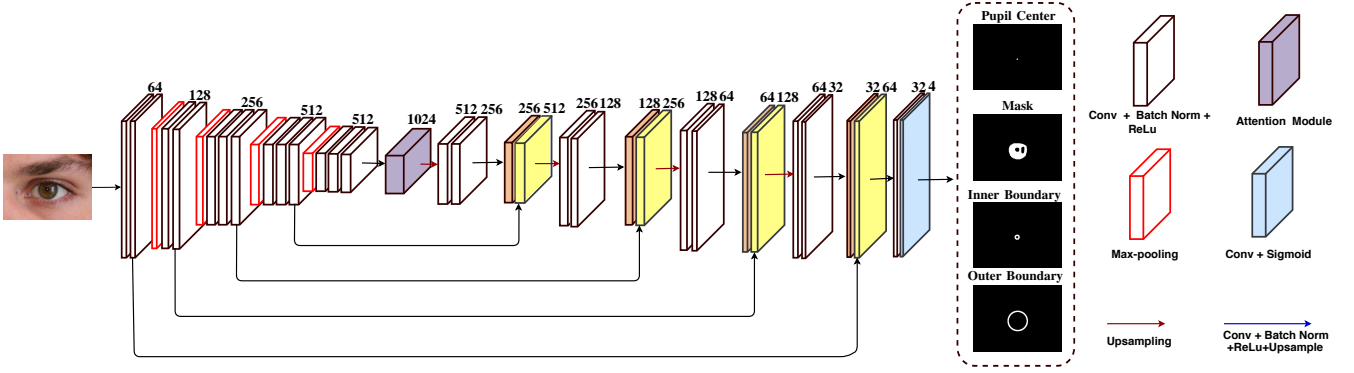
Recently, Fully Convolutional Networks (FCNs) have been widely applied in many tasks such as semantic segmentation\cite{long2015fully,chen2014semantic,chen2016deeplab,chen2017rethinking,ronneberger2015u-net}, edge detection\cite{yu2017casenet} and salient object detection\cite{hou2017deeply}. FCNs are built only with locally connected layers, such as convolution, pooling and upsampling layers, and no dense layers such as fully connected layer are used. Hence, FCNs could take images of arbitrary size as input and produce corresponding-sized output, which is desired in spatially dense prediction tasks.

Accordingly, we propose a multi-task Fully Convolutional Encoder-Decoder Attention Network framework, shown in \figref{fig:arch-network}, which contains an Encoder path and a Decoder path. The Encoder path encodes feature maps of CNN models by convolution, ReLu,~\etc.,~to capture semantic information. The Decoder path decodes the feature maps to recover spatial information lost in the pooling layers by concatenation with feature maps of the Encoder path.

The Encoder path adopts VGG-16\cite{Simonyan2014Very} as the encoding network. We remove the fully convolutional layers and the remaining network is used to learn hierarchical features. The whole encoding network could be divided into 5 stages and every stage is composed of a serial of convolutional layers, batch normalization layers, ReLU layers, and max-pooling layers which gradually reduce the size of feature maps. In lower stages, the feature maps contain more low-level spatial information such as edges but lack semantic information due to small receptive fields. In higher stages, bigger receptive fields extract more semantic information and embed it in the feature maps. In fact, many similar networks, such as ResNet\cite{he2016deep} and DenseNet\cite{huang2017densely}, could also be used as the encoding network.

As described in \cite{zhao2017pyramid}, the size of receptive fields could roughly indicate how much the context information is taken into consideration. For dense prediction task, we need to consider both the local spatial features and global, non-local semantic features. Encouraged by the high performance of DeepLab\cite{chen2017rethinking} and PSPNet\cite{zhao2017pyramid} on semantic segmentation task, we directly adopt atrous spatial pyramid pooling (ASPP) and Pyramid Pooling Module (PSP) for effectively extracting multi-scale receptive fields to reflect multi-scale context information, respectively.

In order to further focus on the most important information and suppress distracting noise, we apply attention mechanism to ASPP and PSP.
Attention mechanism allows us to adjust the weights of different channels in feature maps and also re-estimates the spatial distribution of feature map according to the context\cite{Mnih2014Recurrent,Wang2017Residual,yu2018learning,Woo2018CBAM,Park2018BAM}. Hence, more discriminative features could be learned. Different from \cite{Woo2018CBAM}, we do not apply channel and spatial attention module sequentially, instead, 3D attention maps that integrating cross-channel and spatial information are directly computed.

After the attention module, we gradually up-sample the feature maps to recover the spatial information. Before up-sampling, we need to fuse feature maps from two different layers: the Encoder layer at the same stage and the Decoder layer in the previous stage. The Decoder layer encodes rich context semantic information while the Encoder layer contains the detailed spatial information. The Decoder layer in the previous stage firstly applies two sequential convolutional layers with kernel size of $3\times3$, batch normalization layers and ReLU layers to further refine features and reduce the number of output channels to half of the number of channels of the Encoder layer at the same stage. Then we fuse the two features by element-wise concatenation.

After fusing the feature maps of the final stage, we apply a sequence of $3\times3$ convolutional layer, each followed by a batch normalization layer and a ReLu layer to summarize the final semantic feature. Then, a $1\times1$ convolutional layer with 4 filters and a per-pixel sigmoid function are adopted to generate probability maps of pupil center, iris segmentation mask, iris inner boundary and iris outer boundary.

\subsubsection{ASPP Attention Module}

\label{sec:aspp}
\begin{figure}[!htb]
\centering
  \begin{overpic}[width=\linewidth]{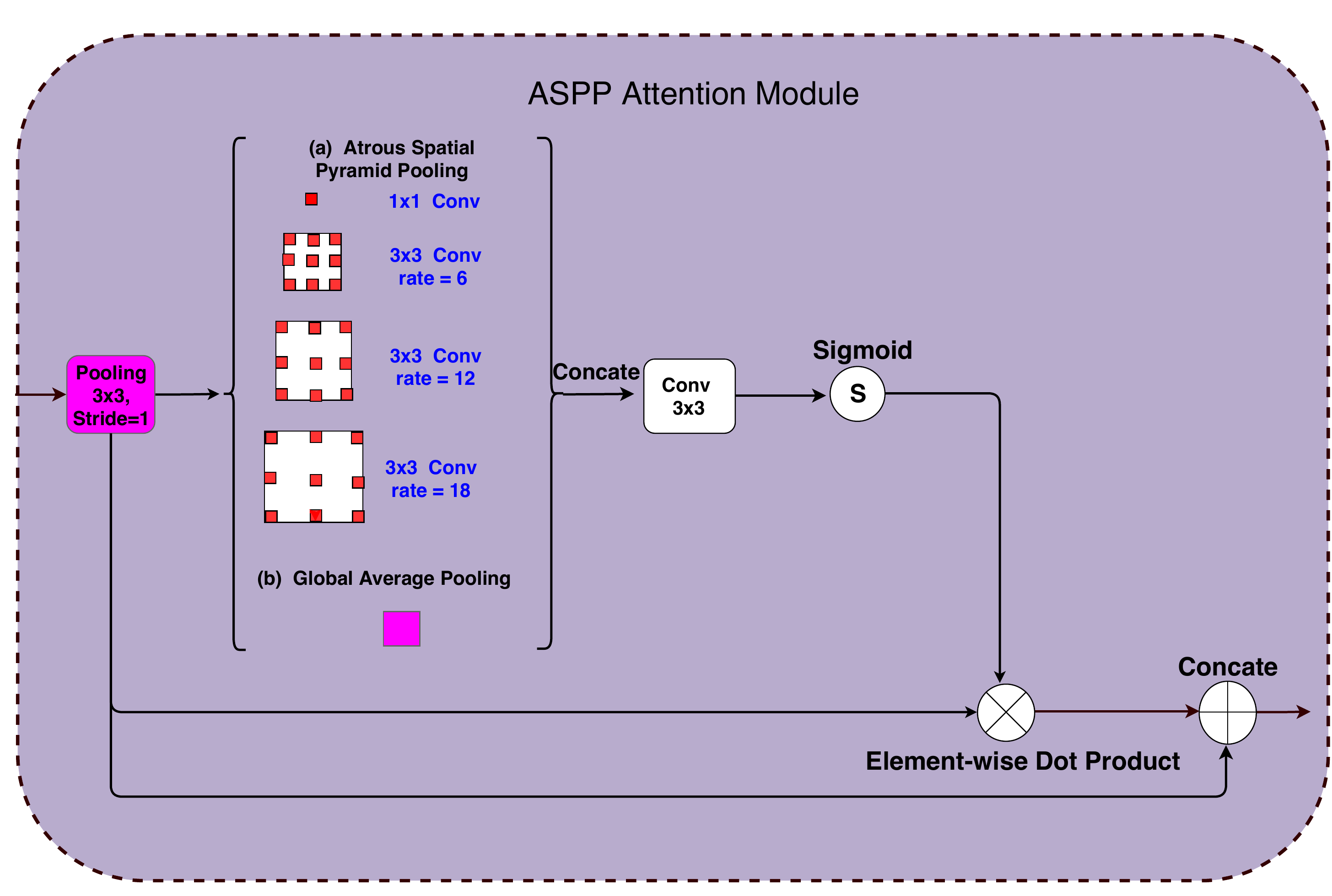}
  \end{overpic}
  \caption{An illustration of ASPP Attention Module. We extract multi-scale context features using multiple parallel filters with different dilation rates along with global average pooling. Afterwards, visual attention map is computed through one single convolution followed by a sigmoid function. Subsequently, the critical regions of input feature map are highlighted by element-wise dot production with obtained attention map. Finally, we concatenate the pooled input feature map before and after attention to get refined features.}
  \label{fig:aspp}
\end{figure}
Atrous Spatial Pyramid Pooling (ASPP) is first proposed in DeepLab V2\cite{chen2016deeplab} which is inspired by the success of spatial pyramid pooling in image classification. In ASPP, dilated convolution (or atrous convolution) with different dilation rates is adopted to extract multi-scale contextual information while keeping the spatial resolution of feature maps unchanged. The original ASPP in DeepLab V2 contains four parallel dilated convolutions with increasing dilation rate, such as 6,12,18,24, on top of the last feature map of the model. In DeepLab V3\cite{chen2017rethinking}, ASPP is improved in three aspects:
(1) batch normalization layer is included for scale adjustment;
(2) $1\times1$ convolution is adopted to replace the degenerated dilated convolution with a higher dilation rate, such as 24;
and (3) global average pooling is connected to the last feature maps of the model to capture the global contextual information.
We will incorporate the improved ASPP with attention module to effectively extract important and discriminative features. The detailed structure of ASPP Attention module is illustrated in \figref{fig:aspp}.

Given an intermediate feature map $F$ as input, a pooling layer with kernel size $3\times3$ and stride 1 is used to get the same sized feature map $P$ as the new input map. Then, five parallel modules are used, including one $1\times1$ convolution with 256 filters (as in \eqref{eq:aspp1}), three dilated convolution with 256 filters and dilation rate set to 6,12,18, respectively (as in \eqref{eq:aspp2}-\eqref{eq:aspp4}), and one global average pooling layer followed by one $1\times1$ convolution with 256 filters and a upsampling layer, mapping the feature map back to the desired dimension (as in \eqref{eq:aspp5}).
It is worth noting that all the convolutional layers are followed by a batch normalization layer and a ReLu layer sequentially. These five modules could be mathematically described as follows:
\begin{align}
 \label{eq:aspp1} D_1(P) &= ReLu(BN(Conv_{1\times1}(P)))\\
\label{eq:aspp2}  D_2(P) &= ReLu(BN(Conv_{3\times3}^{6}(P)))\\
\label{eq:aspp3}  D_3(P) &= ReLu(BN(Conv_{3\times3}^{12}(P)))\\
 \label{eq:aspp4} D_4(P) &= ReLu(BN(Conv_{3\times3}^{18}(P)))\\
 \label{eq:aspp5} G(P) &= Up(ReLu(BN(Conv_{1\times1}({AvgPool(P)}))))
\end{align}
The above feature maps are fused as:
\begin{equation}\label{eq:aspp6}
H= D_1(P)\oplus D_2(P) \oplus D_3(P) \oplus D_4(P) \oplus G(P)
\end{equation}
where $\oplus$ represents channel-wise concatenation. Then, we apply one single $3\times3$ convolution to refine the fused feature maps and reduce the number of output channel to 512 to match with the input feature map $F$. The final 3D attention map $M(F)$ is produced by applying a per-pixel sigmoid operation to refined feature maps. As a result, values of attention map $M(F)$ are bounded in [0,1], where the bigger value indicates the higher importance.

To focus on the more discriminative features of input feature map, the final fusion operation is defined as:
\begin{equation}\label{eq:aspp7}
F'=P\oplus(P\otimes(M(F)))
\end{equation}
where $\otimes$ represents element-wise dot product operation. The above design makes fused feature maps focus only on the most important parts of an input signal. At the same time, the original input is also concatenated to the fused ones to keep other valuable information in the original input signal.
\subsubsection{PSP Attention Module}
\begin{figure}[!htb]
\centering
  \begin{overpic}[width=\linewidth]{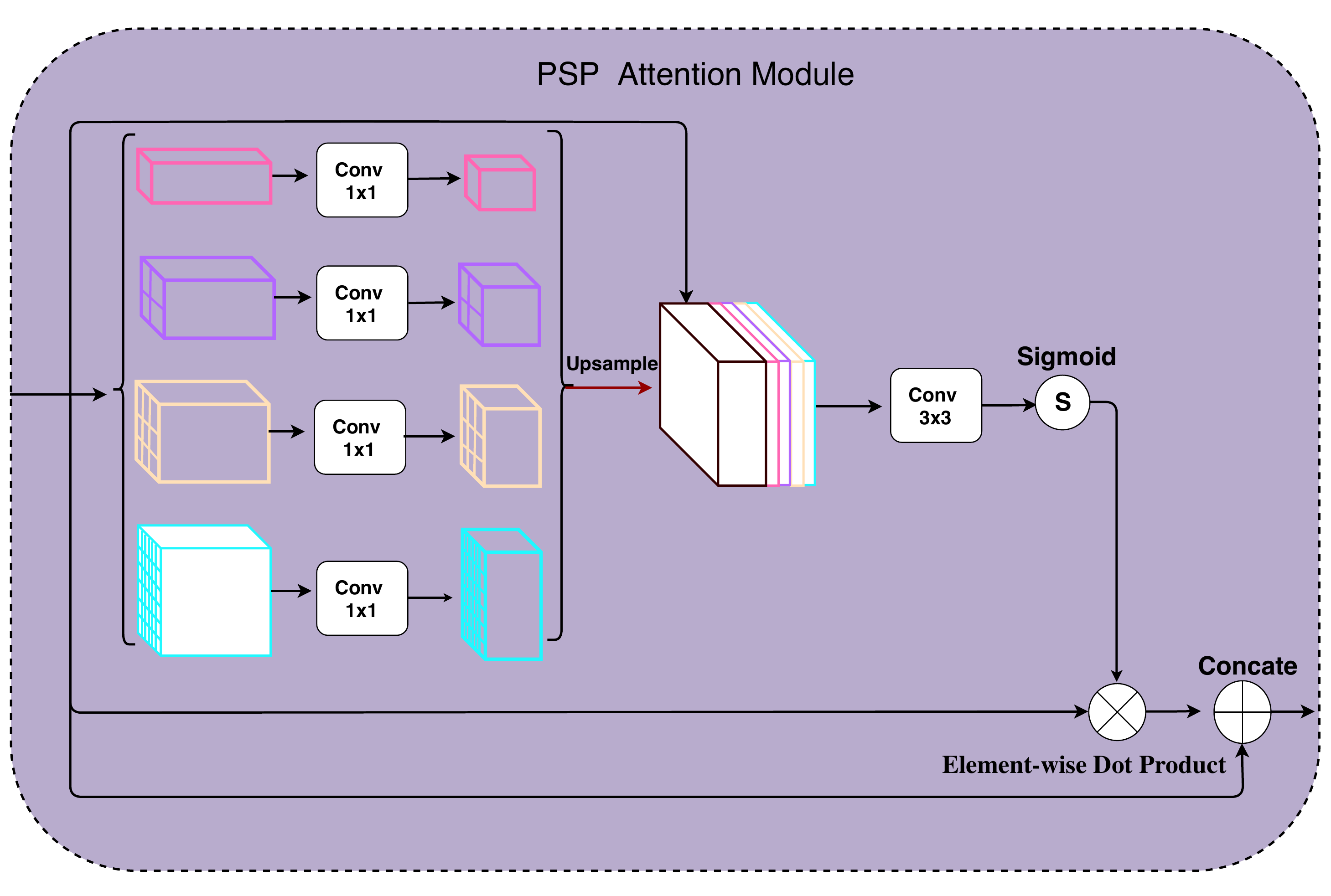}
  \end{overpic}
  %\vspace{6pt}
  \caption{An illustration of PSP Attention Module. We extract both local and global context information by concatenating the input feature map with several sub-region representations of different scales. Then, an attention processing similar to the ASPP Attention module is applied to fused feature maps to get refined features.}
  \label{fig:psp}
\end{figure}
The Pyramid Pooling Module is proposed in PSPNet\cite{zhao2017pyramid} for semantic segmentation. The module fuses multiple features under different pyramid scales which could be controlled by varying bin sizes of pooling. By setting bin sizes to $1\times1$, $2\times2$, $3\times3$ and $6\times6$, an input feature map could be pooled to four different scales. To be concrete, the first pooling operation is actually global average pooling which captures the global contextual information, whereas the other three pooling operations divide the feature maps into different sub-regions and form multi-scale pooled representation for different localizations. Then, a $1\times1$ convolution (and batch normalization, ReLu) is applied to the global and local context representations to reduce the number of output channels to a quarter of the input feature map $F$. To further fuse with original input feature map, we must ensure that the pooled feature maps should have the same resolution as the input feature map. Hence, we upsample the pooled maps to be of the same size as the input feature map via bilinear interpolation. Finally, upsampled feature maps are concatenated with the original input feature map as the final pyramid pooling features $H$. After that, an attention processing similar to ASPP Attention module is applied. The detailed structure of PSP Attention module is illustrated in \figref{fig:psp}.

\subsection{Post-Processing}
\label{sec::post-processing}
Probability maps of pupil center, iris segmentation mask and iris inner/outer boundary could be obtained by forwarding the iris image though the network. Then, we get coarse iris inner/outer contour by using Viterbi algorithm \cite{Sutra2012The} and further fit iris inner/outer circle by using least-squares circle fitting algorithm\cite{chernov2005least}. Before searching the coarse contours, we remove the noise from predicted probability maps and get the range of iris inner/outer circular boundary by a serial of robust image processing operations.

\subsubsection{Edge Denoising \& Boundary Range Estimation}
\label{sec::range}
Different from thin contours produced by traditional edge detection methods such as Canny detector\cite{canny1986computational},~\etc., deep learning based edge detector always produces thick, noisy and blurred edges which are not well aligned to actual image boundaries\cite{xie2015hed}. To eliminate noisy edges, we utilize the prior geometric constraint of pupil center, iris segmentation mask and iris inner/outer boundary and adopt threshold segmentation, connected-component analysis and nearest neighbor search to do the job.

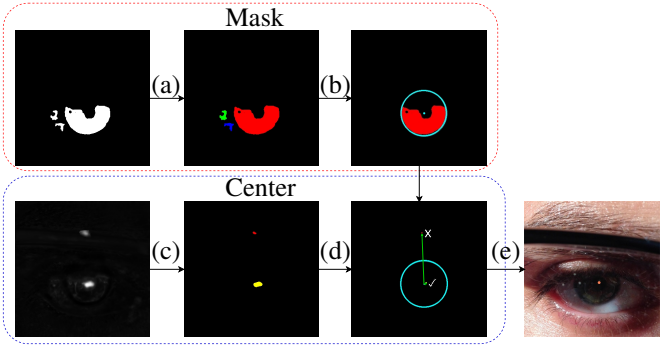
\begin{figure}[!htb]
\centering
  \begin{overpic}[width=\linewidth]{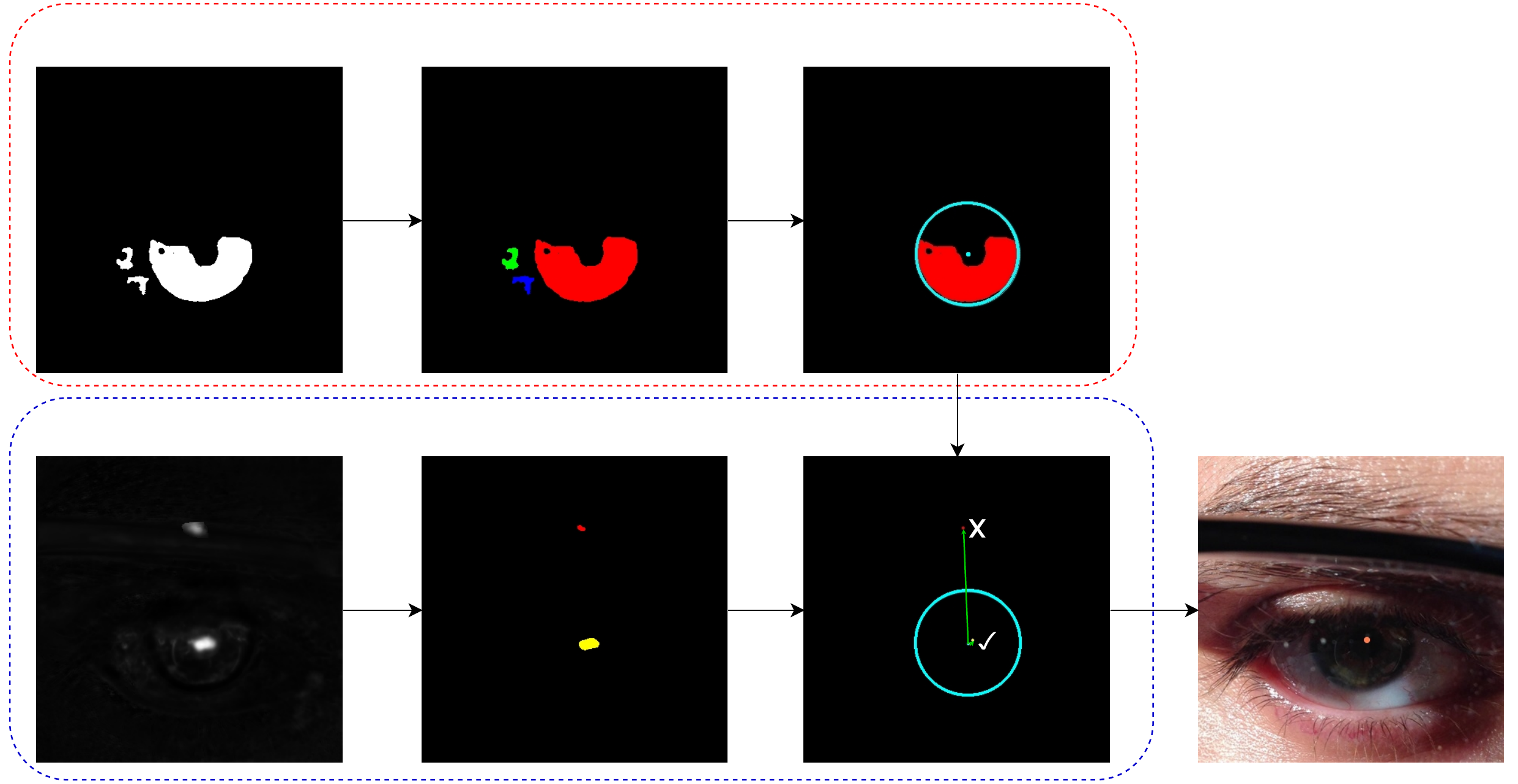}
   \put(34,48){Mask}
   \put(34,22.5){Center}
   \put(21.5,38){ (a)}
   \put(46.8,38){ (b)}
   \put(21.5,13){ (c)}
   \put(46.8,13){ (d)}
    \put(72.5,13){ (e)}
  \end{overpic}
  %\vspace{6pt}
  \caption{Overview of pupil center localization. (a) and (c): threshold segmentation and connected-component analysis; (b):  get the circumcircle of
  max-area mask subregion; (d): nearest neighbor search; (e): get actual pupil center.}
  \label{fig:center}
\end{figure}
To be specific, we locate the pupil center in the first place, as shown in \figref{fig:center}. Among the four outputs of the network, iris segmentation mask is the most accurate and max-area iris mask connected subregion has the highest confidence.
For pupil center localization, the point with the highest score in the probability map of pupil center could be considered as a good initialization.
However, there may be more than one candidate center point with high confidence score for some noisy iris images and the highest score could even be achieved by a noisy pixel.
Therefore, we present a more robust alternative for pupil center localization.
Considering the real pupil center point is adjacent to iris mask, the pupil center is located by searching the nearest pupil center subregion from the circumcircle center of max-area iris mask subregion.
Before searching, the probability map of iris mask is segmented using global threshold (200-255) to get iris mask regions with higher confidence. In addition, the probability map of pupil center is segmented by using lower threshold (150-255) to get more candidate regions.
After that, we compute connected components of pupil center and iris mask, and then perform nearest neighbor search. Once the nearest connected component of pupil center is found, we consider its geometric center as the estimated pupil center.
Since iris center is approximately close to pupil center in most of the cases except for serious deformation, we simply initialize iris center using the coordinates of the pupil center.

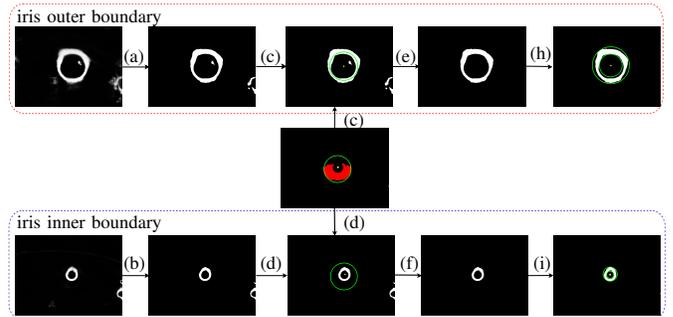
\begin{figure}[!htb]
\centering
  \begin{overpic}[width=\linewidth]{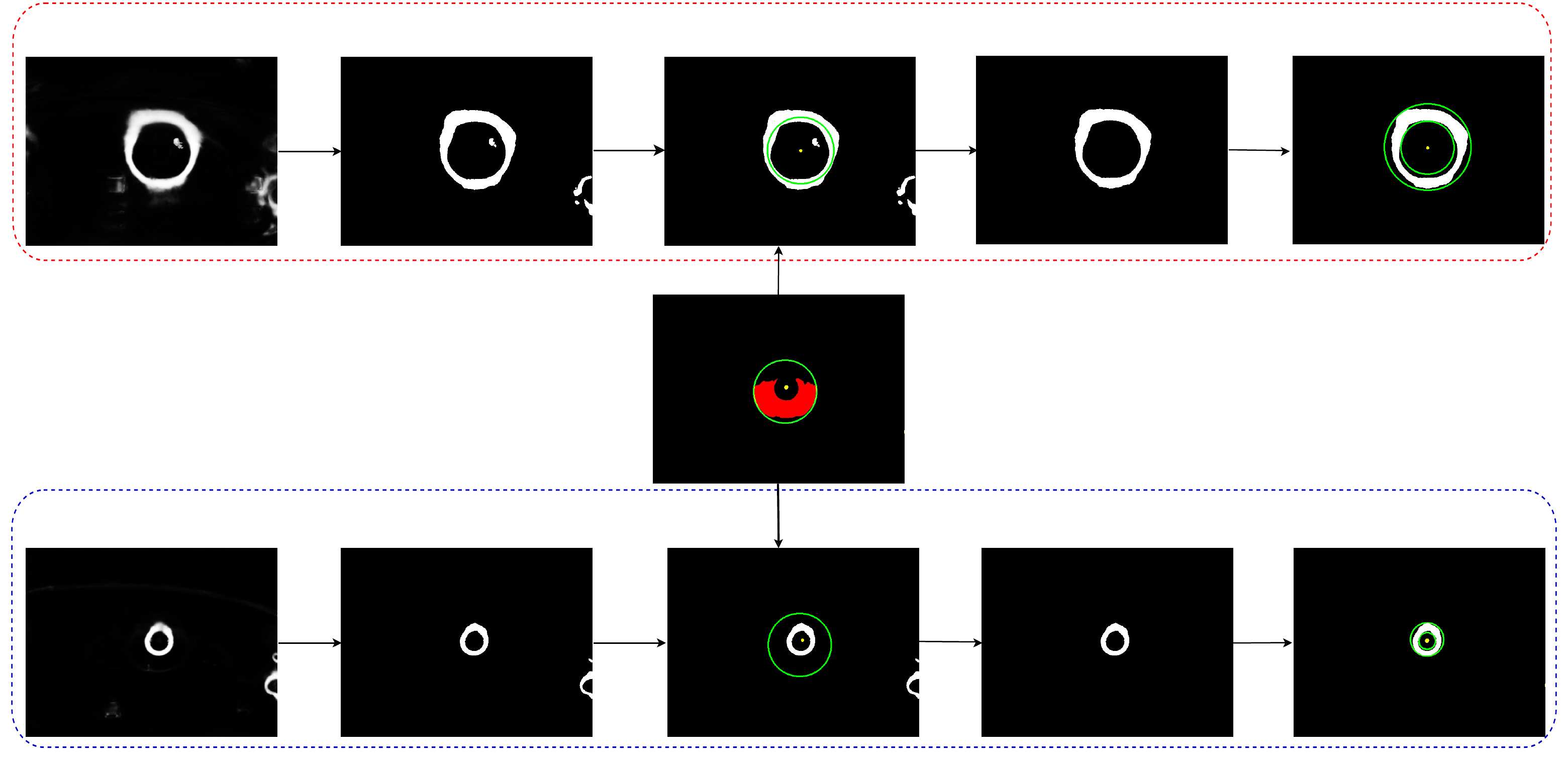}
   \put(2,45.5){\scriptsize iris outer boundary}
   \put(18,39.5){\scriptsize (a)}
   \put(58.8,39.5){\scriptsize (e)}
  \put(79,40){\scriptsize (h)}
   \put(2,14.5){\scriptsize iris inner boundary}
     \put(18,8.5){\scriptsize (b)}
   \put(59.5,8.5){\scriptsize (f)}
  \put(79.5,8.5){\scriptsize (i)}
    \put(51,14.5){\scriptsize (d)}
    \put(51,30){\scriptsize (c)}
    \put(38.5,39.5){\scriptsize (c)}
     \put(38.5,8.5){\scriptsize (d)}
  \end{overpic}%\vspace{-20pt}
  \caption{Edge denoising \& boundary range estimation. (a) and (b): threshold segmentation; (c) and (d): generate target region; (e) and (f): edge denoising; (h) and (i): boundary range estimation.}
  \label{fig:denoise}
\end{figure}
Afterwards, the range of iris inner/outer circular boundary is estimated. Although the majority of noisy edges are removed via applying threshold segmentation, some edges with high-intensity still exist. According to the geometric relationship between iris mask and boundaries, regions where iris boundary is impossible to be located in are further eliminated.
More specifically, an enclosing circle close to actual iris outer boundary is generated by taking estimated pupil center as its origin and the maximum distance between the origin and max-area iris mask as radius. Then, for the iris outer boundary, noisy edges completely falling into the inside and outside of the enclosing circle are excluded. For iris inner boundary, those noisy edges completely falling into the outside of the enclosing circle are also excluded. Finally, we compute the minimum and maximum distances between the pupil/iris center and the refined iris inner/outer boundary. The detailed process is illustrated in \figref{fig:denoise}.

\subsubsection{Iris Inner/Outer Circle Localization}
We modify the original Viterbi algorithm \cite{Sutra2012The} by replacing radial gradient maps with refined probability maps of iris inner and outer boundaries, as well as adopting the estimated range of iris inner/outer circular boundary to output coarse iris inner and outer contours. Then, least-squares circle fitting algorithm\cite{chernov2005least} is applied on coarse contours to estimate the parameters of iris inner/outer circular boundary.

\subsection{Training Objectives}
%%%%%%%%%%%%%%%%%%%%%%%%%%%%%%%%%%%%%%%%%%%%%%%%%%%%%%%%%%%%%%%%%
We optimize all the outputs of IrisParseNet in an end-to-end manner simultaneously. More formally, given an input image $X = \{x_j, j=1,...,|X|\}$ of arbitrary size, we are interested in obtaining probability maps of pupil center, iris segmentation mask, iris inner boundary and iris outer boundary, each of the same size as $X$.

\subsubsection{Pupil Center Detection}
We denote $P = \{p_{j}, j=1,...,|X|\}$ as the predicted probability map of pupil center, in which $p_{j}\in[0,1]$ indicates the probability of pixel $x_j$ being the pupil center, and index $j$ samples every possible spatial location in the input image $X$.

The ground truth of pupil center, denoted by $\bar{P} = \{\bar{p}_{j}, j=1,...,|X|\}$, is a binary image, where pixel value $\bar{p}_{j}$ being 1 suggests that the pixel ${p}_{j}$ belongs to the pupil region, otherwise is part of the background. Due to shortcomings of deep learning models for dense prediction task, the labeled ground truth of pupil center is not a single pixel but a set of pixels located in the neighborhood of the actual pupil center, see \secref{sec:anno}.

Due to the extreme imbalance of the number of positive and negative samples in the result of pupil center detection (most of the pixels are background), we use focal loss\cite{lin2017focal} as the objective function to alleviate this problem. Focal loss introduces two hyper parameters, \ie~$\alpha$ and $\gamma$, to be tuned for better performance:
\begin{equation}
\begin{split}
\mathcal{L}_{\text{pupil}} &= l(P,\bar{P})\\
  &=\sum_{j}\Big[-\alpha(1-\tilde{p}_{j})^{\gamma}\log(\tilde{p}_{j})\Big],
\end{split}
\label{eq:pupil_loss}
\end{equation}
 where
\begin{equation}
  \tilde{p}_{j} =
  \begin{cases}
    p_{j} &\text{if $\bar{p}_{j}=1$}\\
    1-p_{j} &\text{otherwise}.
  \end{cases}
  \label{eq:quant}
\end{equation}

\subsubsection{Iris Segmentation}
Since iris segmentation can be seen as a binary semantic segmentation task, we simply adopt a standard binary cross-entropy loss to supervise the training process.
Let $S = \{s_{j}, j=1,...,|X|\}$ denote the predicted probability map of iris segmentation mask, where $s_{j}$ represents the probability of pixel $x_j$ locating in the iris area.
The corresponding binary ground truth of iris segmentation mask is denoted as  $\bar{S} = \{\bar{s}_{j}, j=1,...,|X|\}$, where $\bar{s}_{j}$ is set to 1 if pixel ${s}_{j}$ is part of the iris region, otherwise $\bar{s}_{j}$ equals to 0. The cross-entropy loss for iris segmentation can be formulated as:
\begin{equation}
\begin{split}
\mathcal{L}_{\text{seg}} &= l(S,\bar{S})\\
  &=\sum_{j}\Big[-\bar{s}_{j}\log(s_{j})-(1-\bar{s}_{j})\log(1-s_{j})\Big],
\end{split}
\label{eq:pupil_loss}
\end{equation}
%\begin{equation}\label{eq:segloss}
%  L_{seg}=\sum_{j}\Big[-\bar{s}_{j}\log(s_{j})-(1-\bar{s}_{j})\log(1-s_{j})\Big]
%\end{equation}

\subsubsection{Iris Inner/Outer Boundary Detection}
Inspired by CASENet\cite{yu2017casenet}, we define iris inner/outer boundary detection as a two-class edge detection problem. To address the problem of positive/negative imbalancing in edge detection, we use the class-balanced cross-entropy loss function which is firstly introduced in HED\cite{xie2015hed}. Suppose the probability maps of iris inner/outer boundary are denoted as $\{E^{1},E^{2}\}$, in which $E^{k}=\{e_{j}^{k}, j=1,...,|X|,k=1,2\}$ and $e_{j}^{k}$ represents the probability of pixel $x_j$ belonging to iris inner boundary ($k=1$) or iris outer boundary ($k=2$). We also manually label the inner and outer boundaries for each iris image, and the ground-truth boundaries are denoted as $\{\bar{E}^{1},\bar{E}^{2}\}$, where $\bar{E}^{k}=\{\bar{e}_{j}^{k}, j=1,...,|X|,k=1,2\}$ is a binary image indicating the distribution of iris boundaries.  The class-balanced cross-entropy loss is formulated as:

\begin{equation}
\begin{split}
\mathcal{L}_{\text{edge}}&=l(E^{1},E^{2};\bar{E}^{1},\bar{E}^{2})\\
  &=\sum_k \sum_j \Big[ -\beta \bar{e}_{j}^{k}\log(e_{j}^{k})\\
  &~~~~-(1-\beta)(1-\bar{e}_{j}^{k})\log(1-e_{j}^{k})\Big],
\end{split}
\label{eq:edgeloss}
\end{equation}
where $\beta$ is the percentage of non-edge pixels in the iris image.

The overall loss function can be expressed as follow:
\begin{equation}
\begin{split}
\mathcal{L} \big( h(X|W), G \big) &=\lambda_1 \mathcal{L}_{\text{pupil}} + \lambda_2 \mathcal{L}_{\text{seg}} + \lambda_3 \mathcal{L}_{\text{edge}} \\
  &= \lambda_1 l(P,\bar{P}) + \lambda_2 l(S,\bar{S}) + \lambda_3 l(E^{1},E^{2};\bar{E}^{1},\bar{E}^{2})
\end{split}
\label{eq:allloss}
\end{equation}
where $\{P,S,E^{1},E^{2}\}=h(X|W)$ is the prediction from IrisParseNet, $G=\{\bar{P},\bar{S},\bar{E}^{1},\bar{E}^{2}\}$ is the corresponding ground truth. $h(X|W)$ is the model hypothesis taking image $X$ as input, parameterized by $W$. We can obtain the optimal parameters by minimizing the overall loss function as follow:
\begin{equation}\label{eq:minloss}
 (W)^* = \argmin \mathcal{L}.
\end{equation}

The hyper-parameters $\alpha$, $\gamma$, $\lambda_1$, $\lambda_2$ and $\lambda_3$  are set to 0.95, 2, 10, 1, 1 in our experiments, respectively .

\section{Datasets and Annotation Methods}
\label{sec:dataset}
In this section, we present detailed descriptions of three challenging and popular datasets: CASIA-Iris-Distance\cite{casiav4}, UBIRIS.v2\cite{UBIRISv2} and MICHE-I\cite{miche_dataset} and our annotation methods.

\subsection{Datasets}
\label{sec::dataset}
\begin{figure}[!t]
  \begin{overpic}[width=1\linewidth]{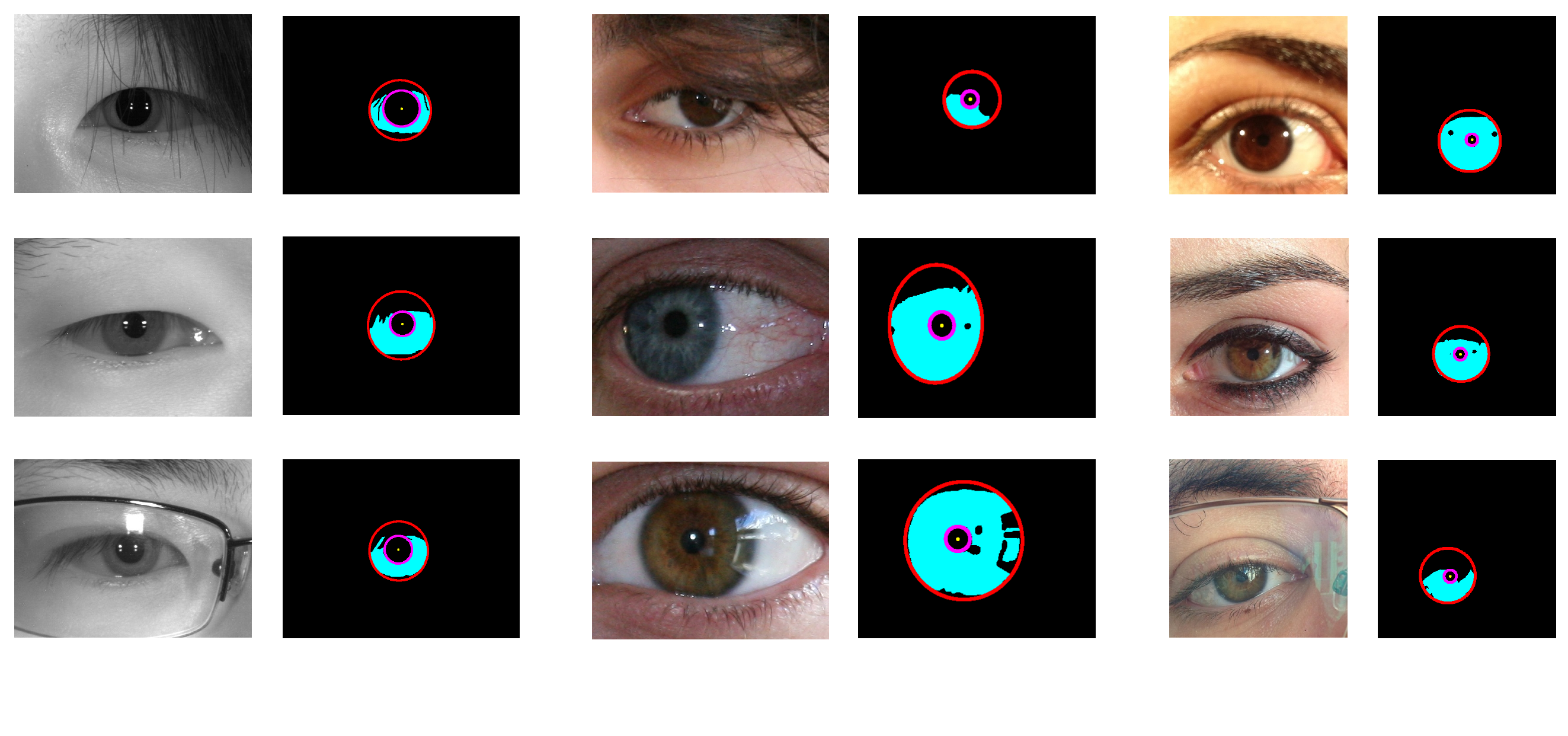}
  \put(3,3){\scriptsize{(a) CASIA-Iris-Distance}}
  \put(45,3){\scriptsize{(b) UBIRIS.v2}}
  \put(79,3){\scriptsize{(c) MICHE-I}}
  \end{overpic}
  \vspace{-13pt}
  \caption{Example images and corresponding ground truths
    (including iris center(chartreuse), iris inner boundary(magenta), iris outer boundary(red), iris segmentation mask(aqua)) of three iris datesets. Best viewed in color.}
  \label{fig:iris_gt}
  \vspace{-10pt}
\end{figure}

\begin{enumerate}
\item  \textbf{\emph{CASIA-Iris-Distance (CASIA)}} contains 2576 images from 142 subjects with resolution of 2352$\times$1728 pixels. The images are captured by self-developed cameras and sample iris images are shown in the first column of \figref{fig:iris_gt} (a). In this dataset, iris images are captured from a distance of more than 3 metres under near infrared illumination (NIR) and meanwhile the subject is moving. A subset which was manually labeled by the author\cite{liu2016accurate} is selected. This subset includes 400 iris images from the first 40 subjects and to speed up processing, images are resized to 640$\times$480 pixels. We follow the same settings as in \cite{liu2016accurate} to select first 300 images from the first 30 subjects for training, and the last 100 images from the last 10 subjects are left for testing in the experiments.
\item  \textbf{\emph{UBIRIS.v2 (UBIRIS)}} consists of 11102 images from 261 subjects which are acquired under visible light illumination (VIS). Images in this dataset are captured on-the-move and at-a-distance with Canon EOS 5D camera and involve realistic noises, such as illumination variance, motion/defocus blur and occlusion of glasses and eyelids. In NICE. I competition, a subset of 1000 UBIRIS.v2 images was used. All images were resized to 400$\times$300 pixels and their segmentation ground truths were manually annotated. According to the protocol of NICE.I competition, 500 images are selected for training and another disjoint testing set of 500 images are used for testing. However, the testing set provided by the organizers of the NICE.I competition has only 445 images. The first column of \figref{fig:iris_gt} (b) shows some examples of images in UBIRIS.v2.
\item  \textbf{\emph{MICHE-I (MICHE)}} dataset was created to evaluate and develop algorithms for colour iris images captured by mobile devices. Images in MICHE-I were captured by three mobile devices including iPhone5 (abbreviated IP5, 1262 images), Samsung Galaxy S4 (abbreviated GS4, 1297 images), and Samsung Galaxy Tab2 (abbreviated GT2, 632 images) in uncontrolled conditions with visible light illumination (VIS) and without the assistance of any operator\cite{Marsico2017Results}. Following by \cite{Hu2015Improving},
    140 images are selected for training and another 429 images are used for testing. Besides, we also use the manually labeled segmentation ground truths provided by \cite{Hu2015Improving}. To speed up processing and preserve the aspect ratio, the width of all iris images is resized to 400 and height is resized to maintain the same proportions as the original image. Finally, the size of resized image is approximately $400\times400$. The first column of \figref{fig:iris_gt} (c) shows some examples of iris images in MICHE-I.
\end{enumerate}

The images from these adopted datasets were acquired under different types of less-constrained environments, thus various kinds of noises are taken into consideration. In addition, the imaging light source contains near infrared light and visible light. In summary, these datasets are representative in a variety of iris recognition applications, so it is convicing and reasonable to evaluate the performance of the proposed method using these datasets.

\subsection{Annotation Methods} %Label Generation
\label{sec:anno}
Training the proposed model requires ground truths of iris segmentation, iris inner/outer boundary and pupil center.
Since the ground truth of iris segmentation has already been provided by other literatures, we only need to obtain annotations of the other three objects.
In the whole labeling process, we use the interactive development environment (HDevelop) provided by the machine vision software,~\ie~MVTec Halcon\cite{halcon201211}, which significantly facilitates our annotation work.

We firstly load iris images in a sequence, and then locate the iris inner and outer boundaries by positioning two ellipses close to explicit iris inner and outer boundaries as the ground truth. After that, the center of iris inner elliptical boundary is regarded as the pupil center.

The initially labeled ground-truth boundaries are too thin, with the width equals to one pixel, but the predicted boundaries from deep models are rather thick. The same problem also occurs in pupil center detection. To tackle this inconsistency, inspired by \cite{liu2016learning}, ground-truth images of training set are dilated using morphologic dilation operator with a circular structuring element of radius 3.

Some examples of manually labeled ground truths can be seen in the second column of \figref{fig:iris_gt} (a), (b), (c), which are sampled from CASIA-Iris-Distance, UBIRIS.v2 and MICHE-I iris datasets, respectively. We sought to accurately locate the iris inner and outer boundaries as well as
eliminate all noise presentd to separate the actual iris pixels.

\section{Experiments and Analysis}
In this section, extensive experiments are conducted on three manually annotated datasets mentioned as in \secref{sec:dataset} to evaluate the proposed model.
The implementation details and data augmentation methods are firstly demonstrated, and then the evaluation protocols are described.
Subsequently, the comparisons of our approach with state-of-the-art iris segmentation and localization methods are presented.
Finally, we analyze the contribution of each individual module of the proposed model by ablation study.

\subsection{Implementation Details}
We implement the proposed architecture based on the publicly available \emph{caffe}~\cite{jia2014caffe} framework and the whole network is initialized using the VGG-16 model\cite{russakovsky2015imagenet} pretrained on ImageNet.
We train the network using mini-batch stochastic gradient descent (SGD)\cite{krizhevsky2012imagenet} with batch size of 4, momentum of 0.9 and weight decay of 0.0005.
Inspired by \cite{chen2014semantic}, we use the "poly" learning rate policy where the learning rate is multiplied by $(1-\frac{iter}{max\_iter})^{power}$ with $power$ set to 0.9, initial learning rate set to $1e^{-3}$ and maximal iteration of 30000. All experiments are conducted on a NVIDIA TITAN Xp GPU with 12GB memory and an Intel(R) Core(TM) i7-6700 CPU.

Data augmentation is a simple yet effective way to enrich training data.
During training, we augment training data with random combination of different geometric transformations (scaling, translation, flip, rotation, cropping) and image variations (blur) on-the-fly. Detailed augmentation operations are:
(1) shuffle images (and gt maps) when reaching the end of an epoch;
(2) randomly resize images (and gt maps) to 7 scales (0.5, 0.75, 1, 1.25, 1.5, 1.75, 2.0);
(3) randomly blur images (mean filter, gaussian blur, median blur, bilateral filter, box blur);
(4) randomly translate images (and gt maps) in x and y axis by a uniform factor between -30 and 30;
(5) randomly left or right flip images (and gt maps);
(6) randomly rotate images (and gt maps) by a uniform factor between -60 and 60;
and (7) random crop images (and gt maps) to a fixed size (321 $\times$ 321) at last.
For testing, we drop all augmentation operations and directly apply the model on the original image.

\subsection{Evaluation Protocols}
\label{sec::evaluation}
To quantitatively evaluate the proposed method, we introduce several evaluation protocols for iris segmentation, iris inner/outer circle localization and iris recognition. The details are described as follows:
\begin{enumerate}
  \item Iris segmentation: The NICE. I competition\cite{nice1} provides two metrics to evaluate the accuracy of iris segmentation. The first measurement is the average segmentation error rate, which could be formulated as follows:
\begin{equation}
E1=\frac{1}{n\times c\times r}\sum_{c'}\sum_{r'}G(c', r')\otimes M(c', r')
\end{equation}
   where $n$ is the number of test images of $r$ rows and $c$ columns. In addition, $G$ and $M$ are the ground truth mask and the predicted iris mask, respectively, and $c', r'$ are the column and row coordinates of pixels in G and M. The operator $\otimes$ represents the XOR operation to evaluate the inconsistent pixels between $G$ and $M$.

\quad The second error measure aims to compensate the disproportion between the apriori probabilities of "iris" and "non-iris" pixels in the images. To be specific, it averages the false positives (fp) and false negatives (fn) rates as follows:
\begin{equation}
E2=\frac{1}{2 \times n}\sum_i(fp+fn)
\end{equation}
  where n is the number of testing images.

\quad We also report the following F-Measure (F1) (the harmonic mean of precision and recall)\cite{hofbauer2014ground} and mean Intersection over Union (mIOU) to provide a comprehensive analysis of the propose method.

\quad The values of E1 and E2 are bounded in $[0,1]$, where the smaller value indicates the better result. Values of F1 and mIOU also fall in the same interval, but the greater value suggests the higher performance in these cases.

  \item Iris inner/outer circle localization: Inspired by \cite{sirinukunwattana2017gland}, we compute the Hausdorff distance between detected iris inner/outer circle (denoted as $D$) and labeled iris inner/outer boundary (denoted as $G$) to measure the shape similarity, which could be defined as:
      \begin{equation}\label{eq:localization}
        H(G,D)=\max\{\sup_{x\in G}\inf_{y\in D}\parallel x-y \parallel, \sup_{y\in D}\inf_{x\in G} \parallel x-y\parallel\}
      \end{equation}
      \quad Smaller Hausdorff distances correspond to higher shape similarity between detected circles and ground truths, suggesting higher detection accuracy.  We report the mean Hausdorff distance (mHdis) for iris inner circle and outer circle to evaluate the performance of localization.
      The average value of the two mean Hausdorff distances demonstrates the overall accuracy of iris localization, thus we include it in the evaluation protocol.

      \quad Besides, inspired by \cite{fuhl2016evaluation}, we also report the detection rate with respect to an error threshold given by the Hausdorff distance between detected iris inner/outer circle and ground truths.
  \item Iris recognition: To verify that our iris segmentation and localization framework is able to improve the performance of iris recognition, we conduct iris recognition experiments with all components but iris segmentation and localization methods fixed. We use the equal error rate (EER) and Daugman's decidability index (DI)\cite{Daugman_id} to quantitatively evaluate the performance of iris recognition. Higher DI values correspond to better discriminative ability of iris recognition systems, meanwhile the iris recognition system with the lowest EER is considered the most accurate.
\end{enumerate}

\subsection{Method Comparison}
\subsubsection{Benchmarks}
We select four representative iris segmentation and localization approaches, including both traditional methods and deep learning based methods, as the benchmark.
In particular, T. Tan \etal\cite{tan2010efficient} proposed an efficient and robust segmentation method to deal with noisy iris images and it could be roughly divided into four processes: clustering based coarse iris localization, pupillary and limbic boundary localization based on a novel integrodifferential
constellation, eyelid localization and eyelash/shadow detection.
The method was ranked the first place in NICE.I competition\cite{nice1}. Since there is no source code available, we only report the result presented in the paper.

RTV-$L^1$\cite{Zhao2015An} proposed a novel total-variation based segmentation framework which used $l^1$ norm regularization to robustly suppress noisy texture pixels to obtain clear iris images. Then, an improved circular Hough transform was used to detect iris and pupil circles on noise-free iris images.
Finally, the authors developed a series of robust post-processing operations to locate iris boundaries more accurately.
We apply the method on above mentioned three datasets using the source code provided by the authors\footnote{The implementation is made available via \url{https://www4.comp.polyu.edu.hk/~csajaykr/tvmiris.htm}}.

Haindl and Krupi$\check{c}$ka\cite{haindl2015unsupervised} proposed an unsupervised segmentation method for colored eye images obtained through mobile devices. The method was ranked \emph{first} in the Mobile Iris Challenge Evaluation (MICHE)-I\cite{de2015mobile} and also outperformed the NICE.I competition winning algorithm, namely T. Tan \etal\cite{tan2010efficient}, with average segmentation error rate $E1$ of 1.24\% on UBIRIS.v2 dataset.
We directly use the executable program\footnote{The executable program is made available via \url{http://biplab.unisa.it/MICHE/MICHE-II/PRL_Haindl_Krupicka.zip}} provided by the authors to test on UBIRIS.v2 and MICHE-I datasets except CASIA-Iris-Distance, as images in CASIA-Iris-Distance are not captured under visible lights.

Besides, MFCNs\cite{liu2016accurate} was the first method that applied fully convolutional network for iris segmentation and achieved better results than previous state-of-the-art methods on CASIA-Iris-Distance and UBIRIS.v2 datasets. We reproduce the method and apply it to our labeled three datasets.

\textbf{Note} that except for RTV-$L^1$, other baseline methods only provide the comparison of iris segmentation mask due to lack of the outputs of iris inner and outer circles.

\subsubsection{Evaluation of Iris Segmentation and Localization}
\label{sec::preprocess_eval}
\begin{figure}[!htb]
  \centering
%\fbox{\rule{0pt}{2in} \rule{0.9\linewidth}{0pt}}
  \begin{overpic}[width=1\linewidth]{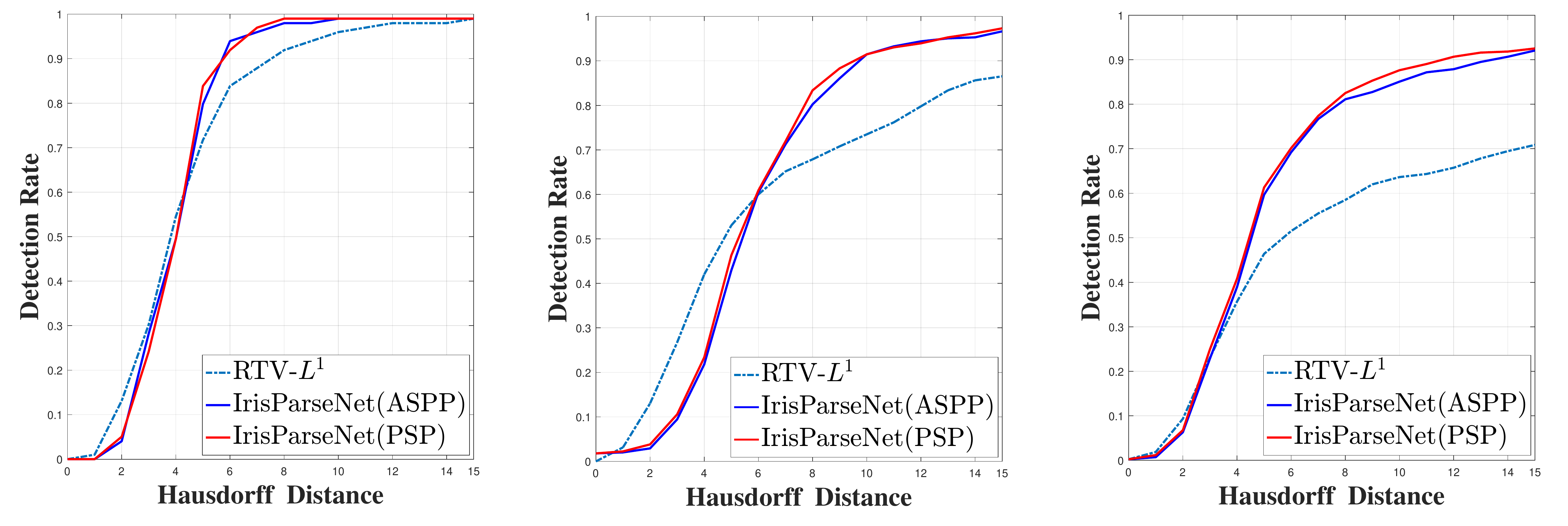}
   \put (8,-3){\tiny(a) CASIA-Iris-Distance}
   \put (44,-3){\tiny(b) UBIRIS.v2}
   \put (79,-3){\tiny(c) MICHE-I}
  \end{overpic}%\vspace{2pt}
  \caption{
   Performance comparison of iris inner circle localization against RTV-$L^1$\cite{Zhao2015An} on the labeled three iris datasets. Success rate is thresholded
   on the Hausdorff distance error.
   }\label{fig:pupil_location}
\end{figure}

\begin{figure}[!hbt]
  \centering
%\fbox{\rule{0pt}{2in} \rule{0.9\linewidth}{0pt}}
  \begin{overpic}[width=1\linewidth]{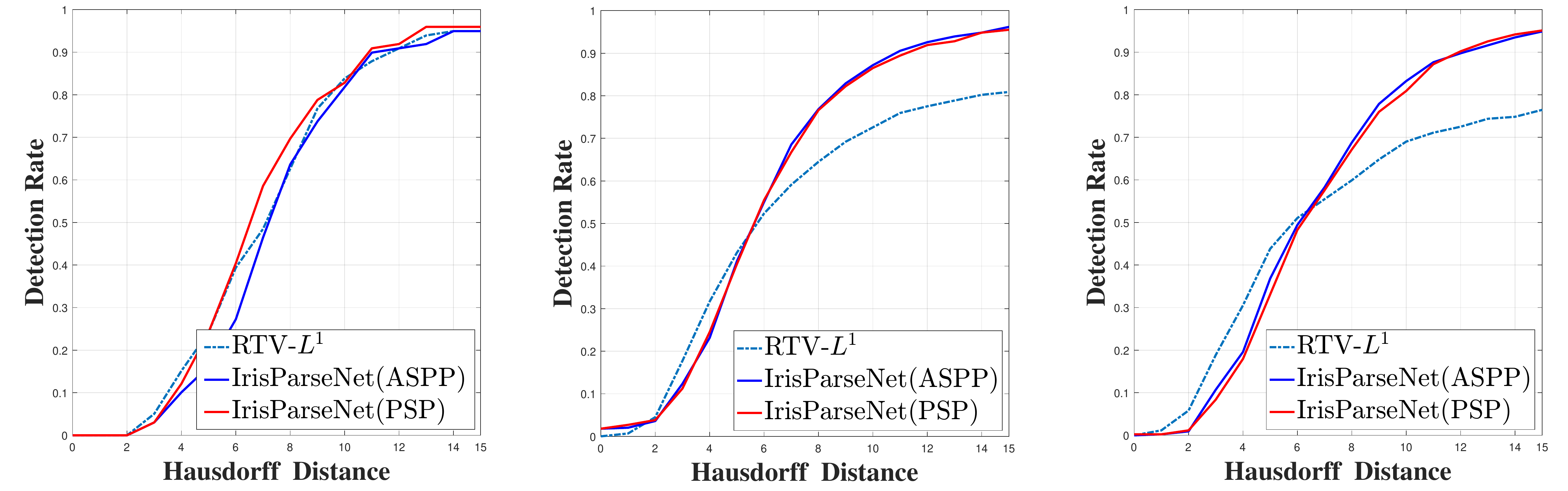}
   \put (8,-3){\tiny(a) CASIA-Iris-Distance}
   \put (44,-3){\tiny(b) UBIRIS.v2}
   \put (79,-3){\tiny(c) MICHE-I}
  \end{overpic}%\vspace{1pt}
  \caption{
   Performance comparison of iris outer circle localization against RTV-$L^1$\cite{Zhao2015An} on the labeled three iris datasets. Success rate is thresholded
   on the Hausdorff distance error.
   }\label{fig:iris_location}
\end{figure}

\tabref{tab:seg} and \tabref{tab:circle}, \figref{fig:pupil_location}, \figref{fig:iris_location} provide summaries of the performance comparison of the proposed method with baseline approaches on iris segmentation and iris inner/outer circle localization under the proposed evaluation protocols.  We also report the storage space of the model and runtime in order to further evaluate the practicability of the proposed method.

As can be seen from \tabref{tab:seg}, IrisParseNet outperforms other approaches on the task of iris segmentation.
Especially, IrisParseNet achieves average segmentation error rates of 0.40\%, 0.84\%, 0.81\% on CASIA-Iris-Distance, UBIRIS.v2 and MICHE-I, respectively.
Hence, our method ranks first according to the NICE. I competition protocol(\textbf{E1}).
Besides, IrisParseNet (including ASPP-type and PSP-type) also achieves better results in terms of mean value (greater than 91\%) and standard deviation (less than 10\%) on F1 metric than other approaches, demonstrating that our approach is highly accurate and robust.
The same superiority is also observed on E2 and mIOU (approximately 85\%) metrics.
The parameters of RTV-$L^1$ are optimized for each dataset, which makes RTV-$L^1$ consistently achieve the good segmentation results on three iris datasets.
It is worth noting that the performance of Haindl and Krupi$\check{c}$ka\cite{haindl2015unsupervised} is not promising, which is inconsistent with the description in their paper.
Although we directly use the execute program provided by the authors when conducting the experiments, we are not able to achieve average segmentation error rates of 1.24\% as described in the original paper for UBIRIS.v2, instead a much higher error rate (3.24\%) is obtained.

From \tabref{tab:circle}, we could see that for the task of iris inner/outer circle localization, IrisParseNet consistently outperforms RTV-$L^1$ on all three datasets under mean Hausdorff distance.
Besides, It could be seen from \figref{fig:pupil_location} and \figref{fig:iris_location} that our method performs comparably to or better than RTV-$L^1$ across the majority of threshold range on all three datasets.

In terms of two types of attention module, IrisParseNet (ASPP) achieves better results on the task of iris segmentation, but IrisParseNet (PSP) shows higher performance on the task of iris inner/outer circle localization.

As for the runtime, the proposed method takes approximate 0.3s, 0.1s, 0.1s for the forward propagation of the network, and 0.4s, 0.4s, 0.4s for post-processing on CASIA-Iris-Distance, UBIRIS.v2 and MICHE-I, respectively.
Compared with traditional approaches, IrisParseNet is more time-efficient (In GPU time), as the overall runtime is less than 0.7s.
Closer observation would reveal that the post-processing step is the most time-comsuming operation, and the runtime of the framework is directly proportional to resolution of input images.

Although our method achieves good segmentation and localization performance, it consumes relative large storage space (approximately 100MB), that limits its application on mobile platforms. To solve this problem, methods such as parameter pruning and sharing, low-rank factorization, knowledge distillation\cite{cheng2017a},~\etc.,~could be adopted to compress the model and further accelerate the training process.

In summary, the proposed IrisParseNet framework demonstrates noticeable superiority over other methods in accuracy, robustness and usability for the task of iris preprocessing.
\begin{table}[!htb]
  \begin{center}
  \renewcommand{\arraystretch}{1.1}
  \setlength\tabcolsep{3pt}
  \caption{Comparison of Different Approaches on the Task of Iris Segmentation Using the Proposed Protocols.}\label{tab:seg}%
  \resizebox{80mm}{80pt}{
  \begin{threeparttable}[t]
  \begin{tabular}{cccccccc}
    \hline
    \multicolumn{1}{c}{\multirow{2}[4]{*}{\bfseries Method}} & \multicolumn{1}{c}{\multirow{2}[4]{*}{\bfseries Dataset}} & \multicolumn{1}{c}{\multirow{2}[4]{*}{\bfseries \tabincell{E1\\(\%)}}} & \multicolumn{1}{c}{\multirow{2}[4]{*}{\bfseries \tabincell{E2\\(\%)}}}
        & \multicolumn{2}{c}{\bfseries F1} &  \multicolumn{1}{c}{\multirow{2}[4]{*}{\bfseries \tabincell{mIOU\\(\%)}}} &\multicolumn{1}{c}{\multirow{2}[4]{*}{\bfseries \tabincell{Average\\Runtime(s)}}} \\
    \cmidrule{5-6}
    &       &       &       &\multicolumn{1}{c}{\bfseries $\mu$(\%)} & \multicolumn{1}{c}{\bfseries $\sigma$(\%)}  &    &  \\
      \hline
   \tabincell{T. Tan \etal\cite{tan2010efficient}}& UBIRIS   &  1.31 & N/A&  N/A & N/A   & N/A &N/A  \\
       \hline
    \multirow{3}{*}{RTV-$L^1$\cite{Zhao2015An}}& CASIA &  0.68 &  0.44  & 87.55 & 4.58 &  78.11 & 2.46  \\
          & UBIRIS   &  1.21 & 0.83  &85.97  & 8.72   & 74.01 & 1.07 \\
          & MICHE    &  2.27 & 1.13  & 77.10 & 14.71  &  64.21 & 1.58\\
       \hline
     \multirow{2}{*}{\tabincell{Haindl and \\Krupi$\check{c}$ka\cite{haindl2015unsupervised}}} & UBIRIS  &  3.24  & 1.62  & 77.03  & 20.67 & 65.08 &  14.33\\
          & MICHE    &  5.08 &  2.54 &  62.19  &  25.28  &  49.79 & 21.94\\
   \hline
     \multirow{3}{*}{MFCNs\cite{liu2016accurate}}& CASIA & 0.50 & 0.25 & 93.14 & 2.97  & 87.30 & 0.47\dag \\
          & UBIRIS   & 0.92 &  0.46  & 90.78   &   4.70 & 81.92 &   0.32\dag\\
          & MICHE    & 0.96   &0.48  & 88.70   &  8.98  & 80.63 &   0.38\dag\\
    \hline
    \multirow{3}{*}{\tabincell{IrisParseNet\\(ASPP)}} & CASIA &\bfseries 0.40 &\bfseries 0.20 &\bfseries 94.30  &\bfseries 3.70  &\bfseries 89.40&    \bfseries 0.25\dag \\
          & UBIRIS   &\bfseries 0.84&\bfseries 0.42 &\bfseries 91.82  &\bfseries 4.26  &\bfseries 85.39  &\bfseries 0.11\dag \\
          & MICHE   & 0.82   &\bfseries 0.41 &91.33 &  8.04  & 84.79 & \bfseries 0.13\dag \\
     \hline
     \multirow{3}{*}{\tabincell{IrisParseNet\\(PSP)}} & CASIA & 0.41 & 0.21 & 94.20  & 3.16  & 89.19 & 0.30\dag\\
          & UBIRIS   &0.85 &\bfseries 0.42 &  91.63 & 4.06 &  85.07  &\bfseries 0.11\dag \\
          & MICHE  &\bfseries 0.81  &\bfseries 0.41 &\bfseries 91.50  &\bfseries 8.01 &\bfseries 85.07 &\bfseries 0.13\dag \\
     \hline
   \end{tabular}
    \begin{tablenotes}
     \item[\dag] GPU time.
   \end{tablenotes}
    \end{threeparttable}}
    \end{center}\vspace{-10pt}
\end{table}

\begin{table}[!htbp]
  \centering
  \renewcommand{\arraystretch}{1.1}
  \setlength\tabcolsep{2pt}
  \caption{Comparison of Different Approaches on the Task of Iris inner/outer circle Localization Using the Proposed Protocols.}\label{tab:circle}%
  \resizebox{80mm}{60pt}{
  \begin{threeparttable}[t]
  \begin{tabular}{ccccccc}
    \hline
   \bfseries Method & \bfseries Dataset & \bfseries \tabincell{mHdis of \\Iris Inner \\Circle} & \bfseries \tabincell{mHdis of \\Iris Outer \\Circle} & \bfseries \tabincell{Overall \\mHdis} &\bfseries \tabincell{Average\\Runtime\\(s)} &\bfseries \tabincell{Overall\\Runtime\\(s)\tnote{1}}\\
      \hline
     \multirow{3}{*}{\tabincell{RTV-$L^1$\cite{Zhao2015An}}}& CASIA &4.24 & 7.74  & 6.08 & N/A & 2.46  \\
          & UBIRIS   &  8.48 & 11.72  &  10.10 & N/A & 1.07  \\
          & MICHE    & 11.96 & 15.49 &  13.73 & N/A   & 1.58 \\
       \hline
     \multirow{3}{*}{\tabincell{IrisParseNet\\(ASPP)}} & CASIA & 4.13 & 7.80 & 5.96 &  0.42\dag &\bfseries 0.67\ddag \\
          & UBIRIS   & 6.06 &\bfseries  6.48 &\bfseries 6.27 & 0.37\dag & 0.49\ddag \\
          & MICHE    & 5.67  &\bfseries 7.33 & \bfseries 6.50  &  0.41\dag & 0.54\ddag \\
     \hline
     \multirow{3}{*}{\tabincell{IrisParseNet\\(PSP)}} & CASIA &\bfseries  4.04 &\bfseries 7.24 & \bfseries 5.64 &\bfseries 0.38\dag & 0.68\ddag \\
          & UBIRIS  &\bfseries 5.99 &6.61 & 6.30 &\bfseries 0.32\dag  &\bfseries 0.43\ddag  \\
          & MICHE   &\bfseries 5.41  & 7.60&\bfseries 6.50 &\bfseries  0.38\dag  &\bfseries 0.51\ddag \\
     \hline
   \end{tabular}
    \begin{tablenotes}
     \item[\dag] GPU time.
     \item[\ddag] GPU time + CPU time.
     \item[1] The overall runtime is the sum of the runtime of iris segmentation and iris inner/outer circle localization.
   \end{tablenotes}
    \end{threeparttable}}%
\end{table}%

\subsubsection{Evaluation of Iris Recognition}
To perform iris recognition (more accurately, iris verification) experiments, we use the full set iris images of CASIA-Iris-Distance, UBIRIS.v2 and MICHE-I datasets. To speed up processing, for CASIA-Iris-Distance and MICHE-I datasets, we use classical Viola-Jones eye detector \cite{Viola2003Rapid} provided by OpenCV to extract the eye region in images, and all eye regions are resized to $400\times 400$. Iris images in UBIRIS.v2 are already scaled to $400\times300$. We use single eye in iris recognition experiments and detailed settings of the experiments are provided in \tabref{tab:setting}.

The proposed IrisParseNet framework is firstly applied for iris segmentation and localization, then Daugman's rubber sheet normalization method\cite{daugman2009iris} is used to produce normalized iris image and iris mask for feature extraction and matching. We adopt the 1-D log Gabor filter to extract iris codes and compute Hamming Distance of iris codes to verify whether two iris are from the same class \footnote{We use the open source iris recognition software package (USIT Version 2) for feature extraction and matching, which is made available via \url{http://www.wavelab.at/sources/Rathgeb16a/}}. The same normalization, feature extraction and matching processes are also adopted in experiments with RTV-$L^1$ \cite{Zhao2015An} and Haindl and Krupi$\check{c}$ka \cite{haindl2015unsupervised}.

Evaluation results of iris recognition are shown in \tabref{tab:recognition}.
From \tabref{tab:recognition}, we could see that experiments using the proposed IrisParseNet framework achieve lower EER and higher DI than those using other methods, especially for CASIA-Iris-Distance, UBIRIS.v2, MICHE-I:iPhone5 and MICHE-I:SamsungGalaxyS4.
Experiment results illustrate that our IrisParseNet method greatly improves the performance of iris recognition.
%模仿booktabs宏包的三线宽度设置 http://bbs.ctex.org/forum.php?mod=viewthread&tid=79220
{%\setlength{\abovetopsep}{0.5ex}
\setlength{\belowrulesep}{0pt}
\setlength{\aboverulesep}{-2pt}
\vspace{-10pt}
\begin{table}[!htb]
  \centering
  \renewcommand{\arraystretch}{1.1}
  \setlength\tabcolsep{5pt}
  \caption{Detailed settings of Iris Recognition Experiment.}\label{tab:setting}%
  \begin{tabular}{c|c|c|c|c|c}
    \hline
    \multirow{2}[4]{*}{\bfseries Dataset} & \multirow{2}[4]{*}{\bfseries CASIA} & \multirow{2}[4]{*}{\bfseries UBIRIS} & \multicolumn{3}{c}{\bfseries MICHE}\\
   \cmidrule(r){4-6}
    &  &  &\bfseries IP5&\bfseries GS4 &\bfseries GT2\\
      \hline
    No. of subjects &  119& 259 & 75 & 75  & 75 \\   \hline
    No. of classes &  238& 518 & 150  & 150& 150 \\   \hline
    No. of images & 2280 &  11100& 995 &764 &438 \\  \hline
    Resolution & $400\times 400$& $400\times 300$&\multicolumn{3}{c}{$400\times 400$}  \\
     \hline
   \end{tabular}\vspace{-15pt}
\end{table}%
}

\begin{table}[!htbp]
  \centering
  \renewcommand{\arraystretch}{1.1}
  \setlength\tabcolsep{5pt}
  \caption{Comparison of Different Approaches on the Task of Iris Recognition Using the Proposed Protocols.}\label{tab:recognition}%
  \resizebox{80mm}{100pt}{
  \begin{tabular}{c|c|c|c}
    \hline
   \bfseries Dataset & \bfseries Method & \bfseries EER & \bfseries DI\\
      \hline
    \multirow{3}{*}{CASIA} & RTV-$L^1$\cite{Zhao2015An}& 0.2708 & 1.1116 \\
     & IrisParseNet (ASPP) &\bfseries 0.0392&\bfseries 3.4474 \\
     & IrisParseNet (PSP) & 0.0412 & 3.4039   \\
     \hline
    \multirow{4}{*}{UBIRIS} & RTV-$L^1$\cite{Zhao2015An}& 0.3303 & 0.9096   \\
     & Haindl and Krupi$\check{c}$ka\cite{haindl2015unsupervised} &0.4249 &0.5069  \\
     & IrisParseNet (ASPP) &0.3107& 0.9642  \\
     & IrisParseNet (PSP) &\bfseries 0.3096  &\bfseries 0.9871   \\
     \hline
       \multirow{4}{*}{MICHE:IP5} & RTV-$L^1$\cite{Zhao2015An}& 0.2279 & 1.3343   \\
     & Haindl and Krupi$\check{c}$ka\cite{haindl2015unsupervised} & 0.3154 & 1.0004  \\
     & IrisParseNet (ASPP) & 0.2045 &\bfseries 1.4994 \\
     & IrisParseNet (PSP) &\bfseries 0.1984 & 1.4896  \\
        \hline
        \multirow{4}{*}{MICHE:GS4} & RTV-$L^1$\cite{Zhao2015An}& 0.2386 & 1.2569   \\
     & Haindl and Krupi$\check{c}$ka\cite{haindl2015unsupervised} &0.3329 & 0.8993  \\
     & IrisParseNet (ASPP) & 0.2038 & 1.3908   \\
     & IrisParseNet (PSP) &\bfseries 0.2029&\bfseries 1.4011   \\
        \hline
        \multirow{4}{*}{MICHE:GT2} & RTV-$L^1$\cite{Zhao2015An}&\bfseries 0.2370 & 1.3509  \\
     & Haindl and Krupi$\check{c}$ka\cite{haindl2015unsupervised} & 0.2948 & 1.1415  \\
     & IrisParseNet (ASPP) & 0.2553&\bfseries 1.3751 \\
     & IrisParseNet (PSP) & 0.2487 & 1.3310   \\
   \hline
   \end{tabular}}\vspace{-10pt}
\end{table}%

\subsection{Ablation Study}
\label{sec:ablation}
We further explore the contribution of each individual module of the proposed model by conducting ablation study.

\subsubsection{Effectiveness of Attention Mechanism}
\begin{figure}[!htb]
  \centering
%\fbox{\rule{0pt}{2in} \rule{0.9\linewidth}{0pt}}
  \begin{overpic}[width=1\linewidth]{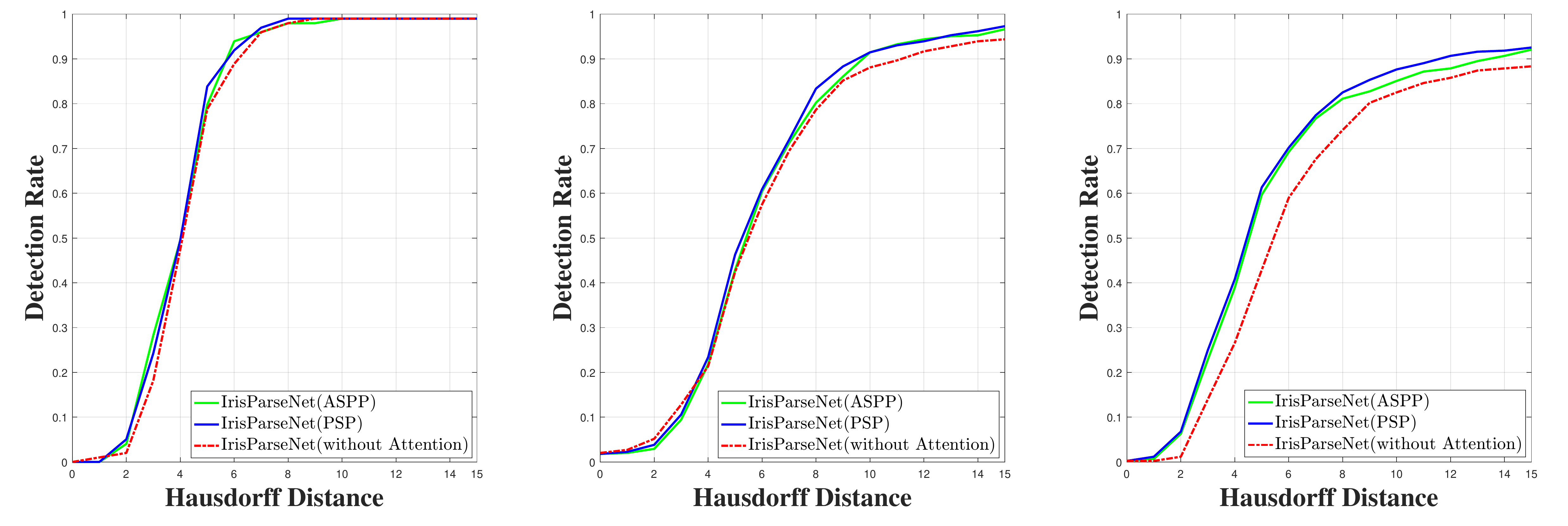}
   \put (7,-3){\tiny(a) CASIA-Iris-Distance}
   \put (46,-3){\tiny(b) UBIRIS.v2}
   \put (80,-3){\tiny(c) MICHE-I}
  \end{overpic} \vspace{2pt}
  \caption{
   Performance comparison of iris inner circle localization with/without attention module on the labeled three iris datasets. Success rate is thresholded
   on the Hausdorff distance error.
   }\label{fig:pupil_attention}\vspace{-4pt}
\end{figure}

\begin{figure}[!htb]
  \centering
  \begin{overpic}[width=1\linewidth]{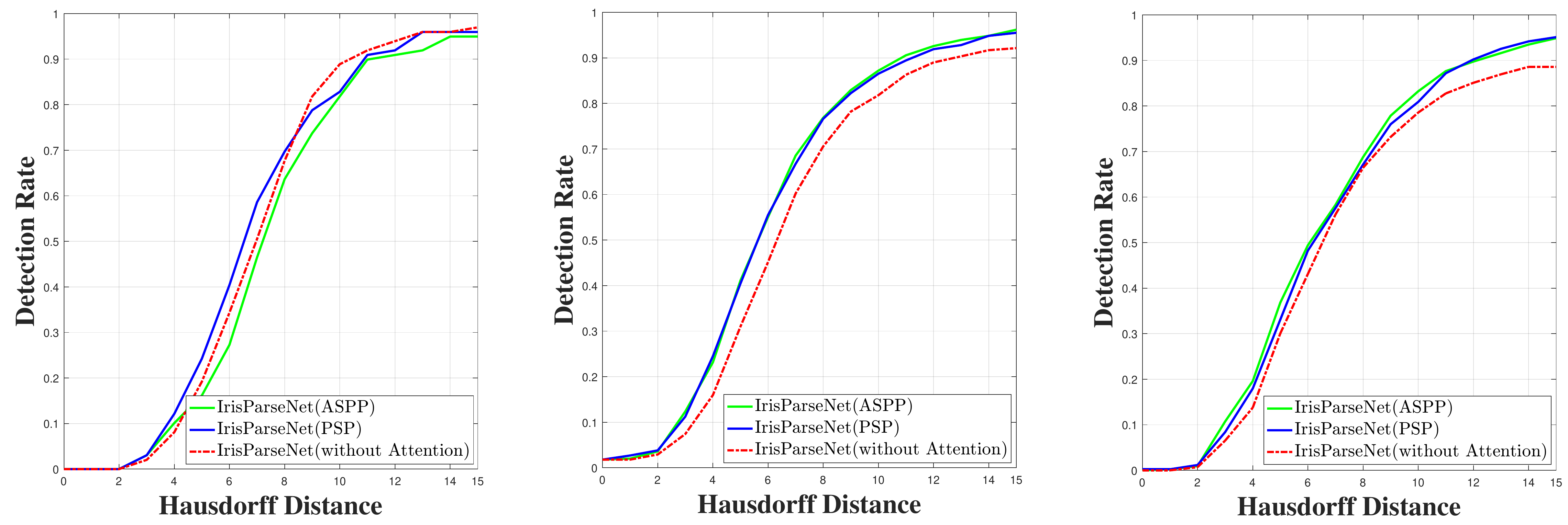}
   \put (7,-3){\tiny(a) CASIA-Iris-Distance}
   \put (46,-3){\tiny(b) UBIRIS.v2}
   \put (80,-3){\tiny(c) MICHE-I}
  \end{overpic} \vspace{2pt}
  \caption{
   Performance comparison of iris outer circle localization with/without attention module on the labeled three iris datasets. Success rate is thresholded
   on the Hausdorff distance error.
   }\label{fig:iris_attention}
\end{figure}

To verify the effectiveness of the attention module, we replace it with two sequential convolutional layers with 256 filters and 512 filters (along with batch normalization layer and ReLU layer).
Experiment results are shown in \tabref{tab:attention}, \figref{fig:pupil_attention}, and \figref{fig:iris_attention}.
From \tabref{tab:attention}, we could see that compared with the original IrisParseNet framework, its variants without attention module suffer from significant performance drop on the task of iris segmentation for all datasets.
As for the task of iris localization, removing attention module would result in a significant performance decrease on UBIRIS.v2 and MICHE-I, as shown in \figref{fig:pupil_attention} and \figref{fig:iris_attention}.

\begin{table}[!htb]
  \centering
  \renewcommand{\arraystretch}{1.3}
  \setlength\tabcolsep{1pt}
  \caption{Comparison of IrisParseNet with/without Attention Module.}\label{tab:attention}%
   \resizebox{80mm}{60pt}{
  \begin{tabular}{ccccccc}
    \hline
   \bfseries Dataset& \bfseries Method  & \bfseries \tabincell{E1\\(\%)} &\bfseries \tabincell{E2\\(\%)}  & \bfseries \tabincell{mean F1\\(\%)} & \bfseries \tabincell{mIOU\\(\%)} & \bfseries \tabincell{Overall mHdis \\of Iris Localization}\\
      \hline
     \multirow{3}{*}{CASIA} & ASPP-type & \bfseries 0.40 & \bfseries 0.20 & \bfseries 94.30  & \bfseries  89.40  & 5.96\\
      & PSP-type &  0.41  & 0.21  &94.20  & 89.19 & \bfseries 5.64\\
       \rowcolor{peach-orange}
      & without Attention  &  0.43 & 0.21  &94.10  & 89.00 & 5.76\\
       \hline
      \multirow{3}{*}{UBIRIS} &ASPP-type & \bfseries  0.84 &  \bfseries 0.42 &\bfseries 91.82 &  \bfseries 85.39 &\bfseries 6.27\\
      & PSP-type & 0.85  &  \bfseries 0.42   &91.63 & 85.07 & 6.30\\
         \rowcolor{peach-orange}
     & without Attention & 0.94  &   0.47 & 90.87 & 83.49 &  7.34\\
     \hline
     \multirow{3}{*}{MICHE}&ASPP-type &  0.82  & \bfseries 0.41 &  91.33 & 84.79 &\bfseries 6.50 \\
      & PSP-type & \bfseries 0.81  & \bfseries 0.41    & \bfseries 91.50  &\bfseries  85.07  & \bfseries 6.50\\
         \rowcolor{peach-orange}
       & without Attention & 0.87  &0.44 & 90.46  & 83.12  & 8.07\\
     \hline
   \end{tabular}}\vspace{-10pt}
\end{table}%

\subsubsection{Effectiveness of Joint Segmentation and Localization}
\begin{figure}[!htb]
  \centering
%\fbox{\rule{0pt}{2in} \rule{0.9\linewidth}{0pt}}
  \begin{overpic}[width=1\linewidth]{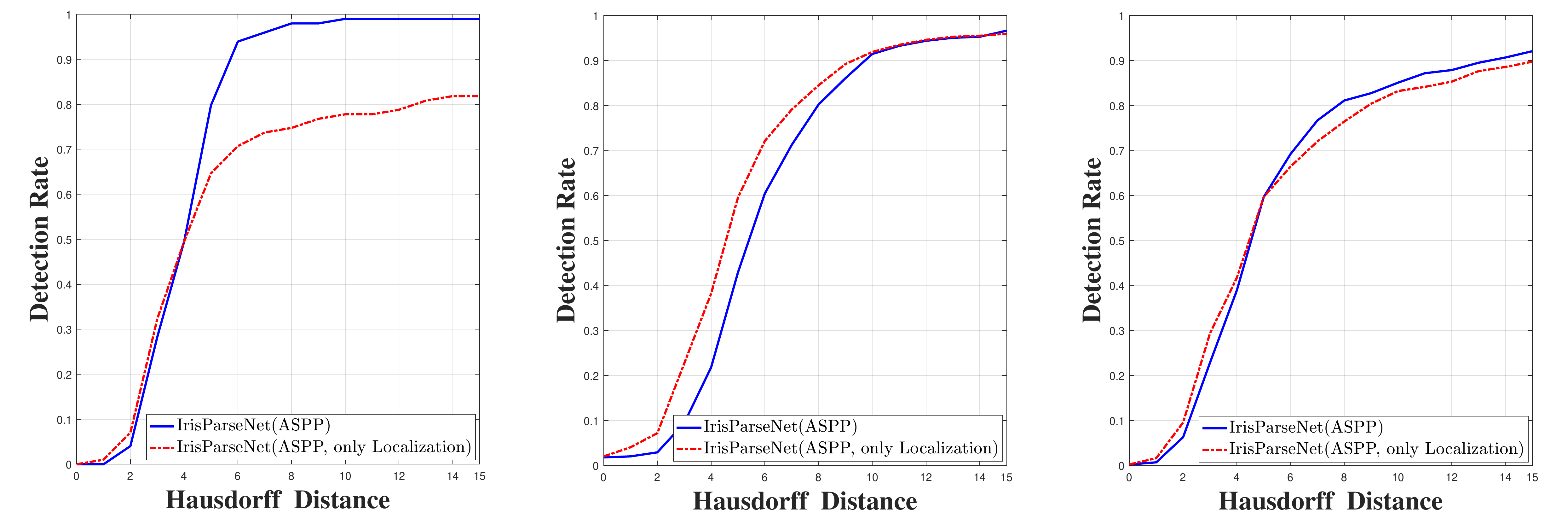}
   \put (8,-3){\tiny(a) CASIA-Iris-Distance}
   \put (44,-3){\tiny(b) UBIRIS.v2}
   \put (80,-3){\tiny(c) MICHE-I}
  \end{overpic} \vspace{2pt}
  \caption{
   Performance comparison of iris inner circle localization with/without joint learning on the labeled three iris datasets. Success rate is thresholded
   on the Hausdorff distance error.
   }\label{fig:pupil_edge}
\end{figure}

\begin{figure}[!htb]
  \centering
  \begin{overpic}[width=1\linewidth]{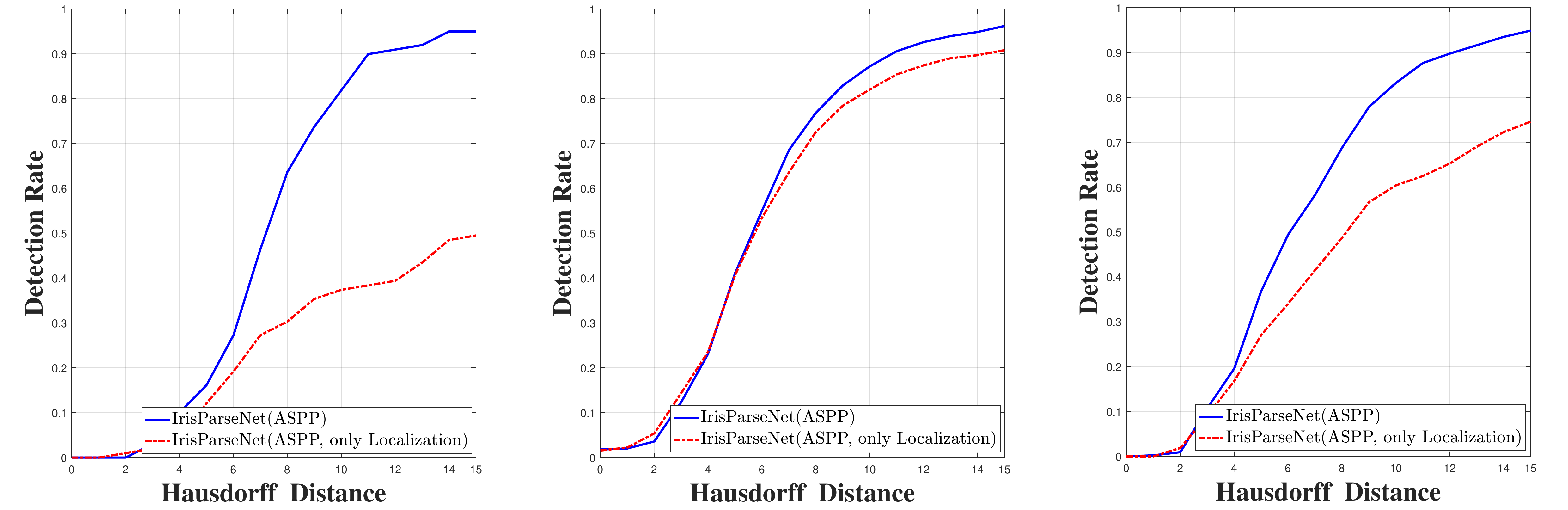}
   \put (8,-3){\tiny(a) CASIA-Iris-Distance}
   \put (44,-3){\tiny(b) UBIRIS.v2}
   \put (80,-3){\tiny(c) MICHE-I}
  \end{overpic} \vspace{2pt}
  \caption{
   Performance comparison of iris outer circle localization with/without joint learning on the labeled three iris datasets. Success rate is thresholded
   on the Hausdorff distance error.
   }\label{fig:iris_edge}
\end{figure}

To evaluate the contribution of joint segmentation and localization, we compare three IrisParseNet framework variants: original IrisParseNet (ASPP), IrisParseNet only with localization part or segmentation part, as shown in \tabref{tab:joint}, \figref{fig:pupil_edge} and \figref{fig:iris_edge}, respectively. Experiment results show that joint learning of iris segmentation and localization helps to improve the performance on both iris segmentation and iris localization tasks.
\begin{table}[!htb]
  \centering
  \renewcommand{\arraystretch}{1.3}
  \setlength\tabcolsep{1pt}
  \caption{Comparison of IrisParseNet with/without Joint training.}\label{tab:joint}%
  \resizebox{80mm}{60pt}{
  \begin{tabular}{ccccccc}
    \hline
   \bfseries Dataset& \bfseries Method  & \bfseries \tabincell{E1\\(\%)} &\bfseries \tabincell{E2\\(\%)}  & \bfseries \tabincell{mean F1\\(\%)} & \bfseries \tabincell{mIOU\\(\%)} & \bfseries \tabincell{Overall mHdis \\of Iris Localization}\\
      \hline
      & ASPP-type & \bfseries 0.40 & \bfseries 0.20 & \bfseries 94.30  & \bfseries  89.40  & \bfseries 5.96\\
      \rowcolor{peach-orange}
     CASIA & only Localization &  N/A   &  N/A   & N/A  &  N/A &  11.91\\
      \rowcolor{light_blue}
      & only Segmentation  & 0.41 & 0.20  & 94.08  & 89.18 & N/A \\
       \hline
       &ASPP-type & \bfseries  0.84 & \bfseries 0.42 &\bfseries 91.82 &\bfseries  85.39 & \bfseries 6.27\\
        \rowcolor{peach-orange}
     UBIRIS & only Localization & N/A &  N/A   &N/A & N/A & 7.39\\
       \rowcolor{light_blue}
     & only Segmentation &  0.85 &   0.42 & 91.70 &  83.37 &  N/A \\
     \hline
        &ASPP-type & \bfseries 0.82  & \bfseries 0.41 & \bfseries 91.33 &\bfseries 84.79 & \bfseries 6.50 \\
         \rowcolor{peach-orange}
     MICHE & only Localization &N/A   & N/A   &N/A  &N/A   &  10.70\\
         \rowcolor{light_blue}
       & only Segmentation & 0.82  &0.41 & 91.32 & 84.72  & N/A \\
     \hline
   \end{tabular}}
\end{table}%

\section{Conclusions and Future Work}
In this paper, we propose a novel deep multi-task learning framework for joint iris segmentation and localization. In this framework, a Fully Convolutional Encoder-Decoder Attention Network and an effective post-processing operation which exploit the priori geometric constraints of pupil, iris and sclera, are proposed to improve the performance of iris segmentation and localization.
Meanwhile, we have collected manual labels of three challenging iris datasets and established comprehensive evaluation protocols, which are publicly available. The proposed method is compared with state-of-the-art methods on the three annotated iris datasets, and shows a leading performance. As for future work, we would explore improving the efficiency of the post-processing step or integrate it into the iris segmentation and localization system to form an end-to-end model.

\bibliographystyle{IEEEtran}
\bibliography{IEEEabrv,IrisSeg}

\end{document}